%% file: main_arxiv.tex
\documentclass[letterpaper]{article} 
\usepackage{aaai2026}  
\usepackage{times}  
\usepackage{helvet}  
\usepackage{courier}  
\usepackage[hyphens]{url}  
\usepackage{graphicx} 
\usepackage{natbib}  
\usepackage{caption} 
\usepackage{algorithm}
\usepackage{subfig}

\usepackage{booktabs}
\usepackage{newfloat}
\usepackage{listings}
\usepackage{wzl_conf}
\usepackage{amsmath}
\usepackage{amssymb} 
\usepackage[noend]{algpseudocode}
\usepackage{colortbl}

\usepackage{tikz}
\usepackage{calc}
\usepackage{multirow}
\usepackage{bbding}
\usepackage{caption}
\usepackage{mathtools}

\usepackage{bm}        
\usepackage{mathrsfs}   
\usepackage{amsfonts}
\usepackage[algo2e,ruled,linesnumbered]{algorithm2e}
\SetKwInOut{Parameter}{parameter}

\usepackage{xcolor}
\usepackage{appendix}

\definecolor{lightgreen}{rgb}{0.9,1,0.9}

\DeclareCaptionStyle{ruled}{labelfont=normalfont,labelsep=colon,strut=off} 
\floatstyle{ruled}
\newfloat{listing}{tb}{lst}{}
\floatname{listing}{Listing}
\title{Realism Control One-step Diffusion for Real-World Image Super-Resolution}

\author {
    Zongliang Wu\textsuperscript{\rm 1, 2, 3, *}, 
    Siming Zheng \textsuperscript{\rm 3, *}, 
    Peng-Tao Jiang\textsuperscript{\rm 3, \#}, 
    Xin Yuan\textsuperscript{\rm 2, \#}  
}

\affiliations {
    \textsuperscript{\rm 1} Zhejiang University, Hangzhou, China \\
\textsuperscript{\rm 2} School of Engineering, Westlake University, Hangzhou, China\\
\textsuperscript{\rm 3} vivo Mobile Communication Co., Ltd\\
\textsuperscript{\rm *} Equal Contribution 
\textsuperscript{\rm \#} Corresponding Author \\
   \{wuzongliang, xyuan\}@westlake.edu.cn, \{zhengsiming, pt.jiang\}@vivo.com
}

\usepackage{bibentry}

\begin{document}

\maketitle

\begin{abstract}
Pre-trained diffusion models have shown great potential in real-world image super-resolution (Real-ISR) tasks by enabling high-resolution reconstructions. While one-step diffusion (OSD) methods significantly improve efficiency compared to traditional multi-step approaches, they still have limitations in balancing fidelity and realism across diverse scenarios. 
Since the OSDs for SR are usually trained or distilled by a single timestep, they lack flexible control mechanisms to adaptively prioritize these competing objectives, which are inherently manageable in multi-step methods through adjusting sampling steps. To address this challenge, we propose a Realism Controlled One-step Diffusion (RCOD) framework for Real-ISR. RCOD provides a latent domain grouping strategy that enables explicit control over fidelity-realism trade-offs during the noise prediction phase with minimal training paradigm modifications and original training data. A degradation-aware sampling strategy is also introduced to align distillation regularization with the grouping strategy and enhance the controlling of trade-offs.
Moreover, a visual prompt injection module is used to replace conventional text prompts with degradation-aware visual tokens, enhancing both restoration accuracy and semantic consistency. 
Our method achieves superior fidelity and perceptual quality while maintaining computational efficiency. Extensive experiments demonstrate that RCOD outperforms state-of-the-art OSD methods quantitatively and visually, with flexible realism control capabilities in the inference stage. Code is available at \url{https://zongliang-wu.github.io/RCOD-SR}.
\end{abstract}


\begin{figure}[t]
\centering
\begin{minipage}[t]{0.46\linewidth}
\centering
\includegraphics[width=1\textwidth ,trim=0 0 0 0, clip]{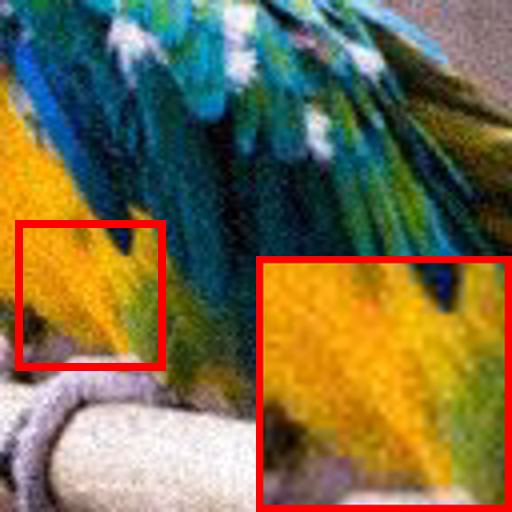}
\parbox{1\textwidth}{\centering\small (a) LR Image \\ ↑PSNR: 19.16, ↓NIQE:10.021}
\end{minipage}
\begin{minipage}[t]{0.46\linewidth}
\centering
\includegraphics[width=1\textwidth ,trim=0 0 0 0, clip]{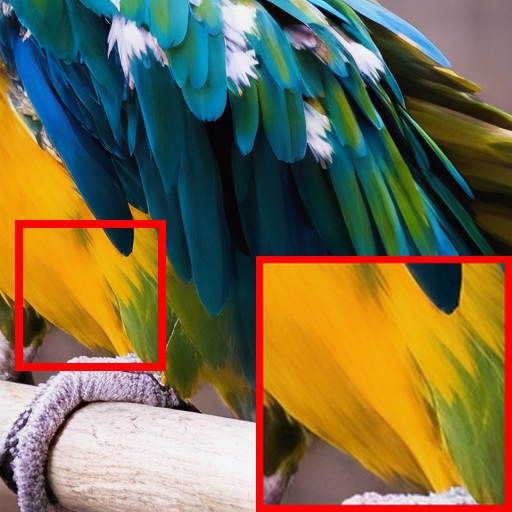}
\parbox{1\textwidth}{\centering\small (d) S3Diff \\ ↑PSNR: 25.47, ↓NIQE: 4.418}
\end{minipage}


\begin{minipage}[t]{0.46\linewidth}
\centering
\includegraphics[width=1\textwidth ,trim=0 0 0 0, clip]{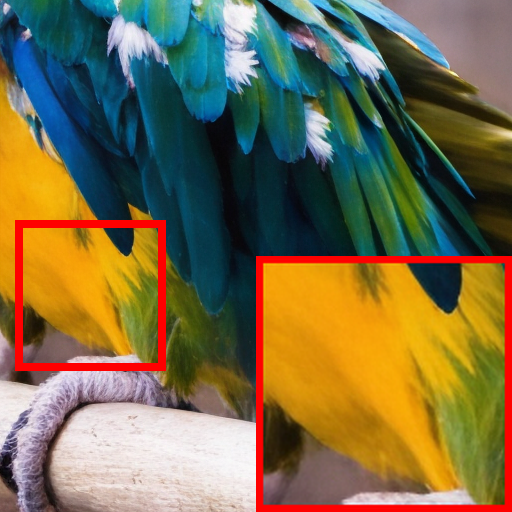}
\parbox{1\textwidth}{\centering\small (e) Ours-Fid. \\ ↑PSNR: \textbf{26.94}, ↓NIQE: 4.049}
\end{minipage}
\begin{minipage}[t]{0.46\linewidth}
\centering
\includegraphics[width=1\textwidth ,trim=0 0 0 0, clip]{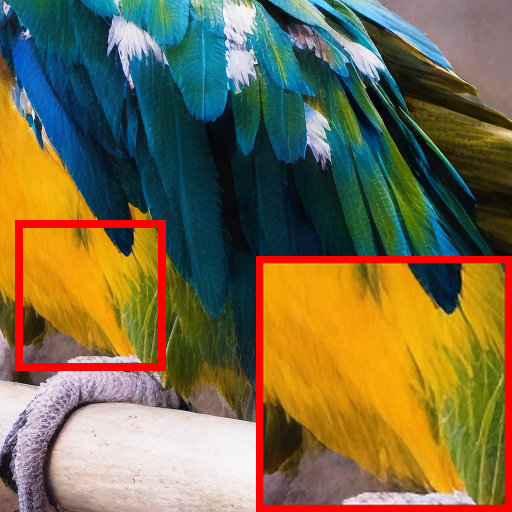}
\parbox{1\textwidth}{\centering\small (f) Ours-Real. \\ ↑PSNR: 25.66, ↓NIQE: \textbf{3.913}}
\end{minipage}
\caption{While previous one-step diffusion methods, such as S3Diff~\cite{zhang2024degradation-s3diff} only yield one optimal result (b), our approach offers the flexibility to control images (c-d) with different fidelity-realism trade-offs during inference, enhancing practical applicability across different scenarios.}
\label{fig:compare_teaser}
\end{figure}

\section{Introduction}
\label{sec:intro}
\begin{figure*}[t]
\centering
\includegraphics[width=1\textwidth, trim=0 80 0 0, clip]{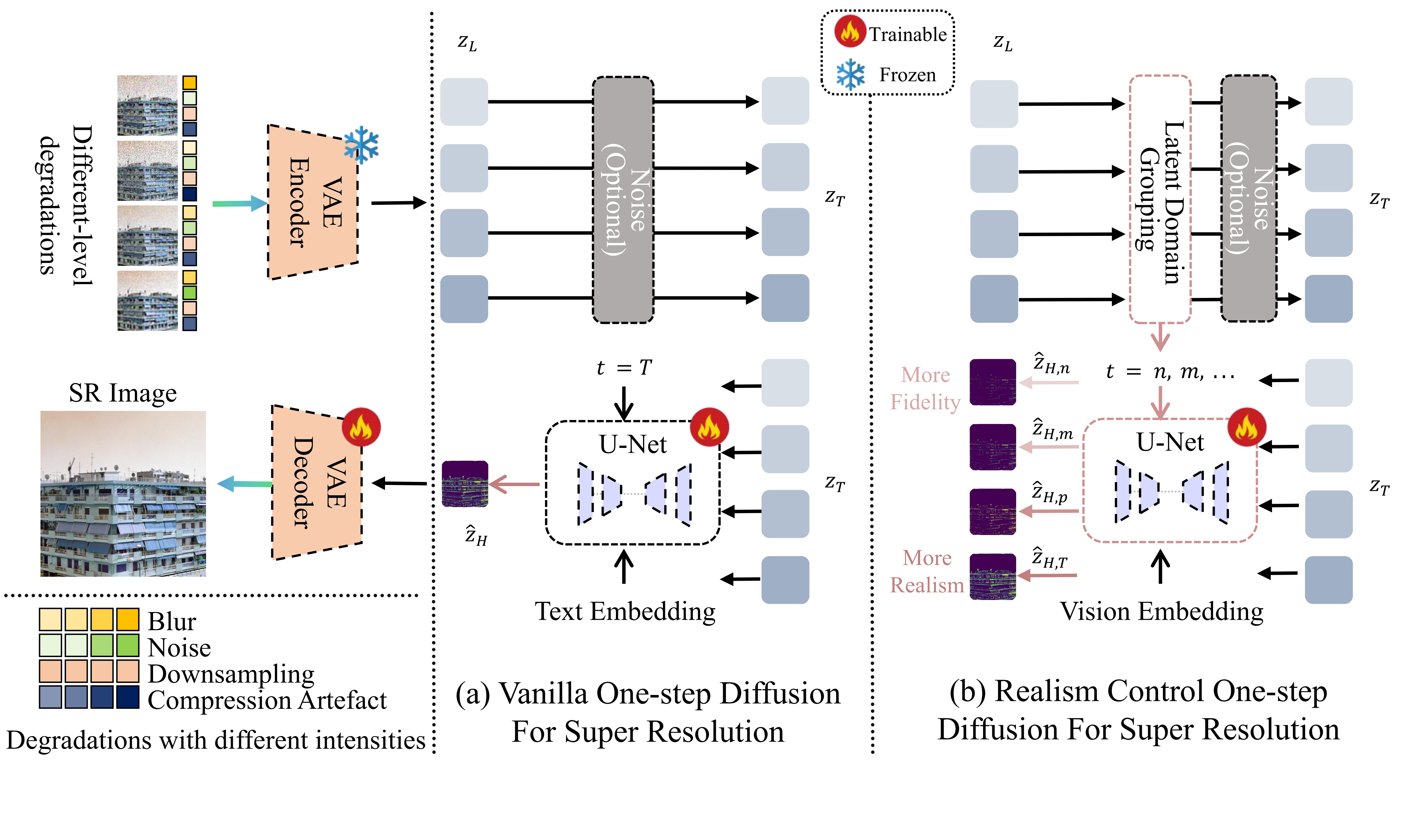}
\caption{Realism control one-step diffusion (RCOD) training process. 
The left part illustrates several synthesized real-world LR images by applying diverse degradations with varying types and intensities on an HR image. 
(a) Existing vanilla one-step diffusion (OSD) methods for super-resolution (SR): These LR images are directly sent into the diffusion forward and reverse process; the denoising U-Net tends to learn to recover the `average' degradation, leading to a monotonous generation ability within the latent domain.
(b) Our proposed Realism Control One-Step Diffusion employs a latent domain grouping strategy. This allows for adaptive control of timesteps (denoising degrees) during the forward process according to the degradation degree in the latent domain. As a result, the denoising U-Net can acquire a more diverse generation capability based on the timestep.}
\label{fig:main_principle}
\end{figure*}

Image super-resolution (SR) \cite{dong2015image, zhang2018image,zhang2021designing,ledig2017photo, liang2021swinir} aims to recover a high-resolution (HR) image from its low-resolution (LR) counterpart. 

Traditional image super-resolution simplifies the degradation process as known noise, blur, or downsampling. In recent years, real-world image super-resolution (Real-ISR) \cite{zhang2021designing,wang2021real} has attracted more attention due to the increasing demand for reconstructing high-resolution images under real-world unknown degradations, which is more challenging and practical in real applications.

While recent advances in Stable Diffusion (SD) models \cite{ho2020denoising,song2020score}, especially the large-scale pretrained text-to-image (T2I) models have demonstrated unprecedented capabilities in various downstream vision tasks \cite{zhang2023adding,rombach2022high}. Some works leveraging pre-trained SD models for multi-step SR, such as DiffBIR~\cite{lin2023diffbir}, StableSR~\cite{wang2024exploiting}, and SeeSR~\cite{wu2024seesr}, have achieved remarkable SR quality through iterative latent space optimization. Though these methods achieve impressive perceptual quality, they suffer from computational inefficiency. The caused latency by multi-step sampling makes real-time applications impractical. 

To address the efficiency concerns, recent attempts focus on one-step diffusion frameworks~\cite{wu2024one,zhang2024degradation-s3diff}, which distill multi-step diffusion priors into single-step inference through knowledge distillation and achieve 10\texttimes to 100\texttimes~speedup over previous multi-step diffusion-based SR methods. However, existing one-step diffusion approaches face a fundamental dilemma: the deterministic single-step generation inherently lacks the controllable fidelity-realism balance that multi-step methods achieve via step-wise noise scheduling. 
As illustrated in Fig.~\ref{fig:compare_teaser}, previous one-step diffusion super-resolution (SR) methods, such as S3Diff~\cite{zhang2024degradation-s3diff}, can only generate a single optimal result but can not meet the need for dynamic adjustment between fidelity and realism. The root cause lies in current OSD training paradigms. Most methods align all the LR inputs under different unknown degradations with a single convergence space through single timestep conditioned training, which results in a balanced static preference for fidelity or realism and prevents adaptive adjustments for scenario-specific requirements. 

Bearing the above concerns in mind, we propose a novel framework that provides one-step diffusion Real-ISR methods with the capability to monotonically control the level of realism. This framework, which we denote as \textbf{Realism Controlled One-step Diffusion (RCOD)}, can be easily integrated into existing one-step diffusion methods for Real-ISR.
Specifically, during the training phase, we incorporate a Latent Domain Grouping (LDG) strategy into the latent diffusion process, grouping training data according to a latent domain metric. Through this strategy, the diffusion denoising network learns to perceive variations in degradation across training samples, thereby gaining adaptive restoration capabilities. Furthermore, to address the inherent limitations caused by text prompts, we introduce a Visual Prompt Injection Module (VPIM) to enhance prompt quality.
Our contributions are summarized as follows:
\begin{itemize}
 
    \item  We propose a simple but effective latent domain grouping (LDG) strategy that {reformulates the noise prediction process by partitioning the latent space into fidelity and realism oriented domains. This allows explicit control over detail preservation versus generative enhancement through simple training paradigm modifications, without requiring additional trainable parameters.}
  \item {During distillation, we introduce a degradation-aware sampling (DAS) strategy that reformulates timestep sampling in the pretrained model by adaptively aligning it with our LDG framework, enhancing controlling with regularization strength.}
    \item To reduce the computational burden of VLM and dependencies on manual text prompts, we propose a visual prompt injection module (VPIM) to replace text prompts with degradation-aware visual tokens, enhancing both restoration accuracy and semantic consistency.
    \item We empirically evaluate our approach on widely used stable diffusion-based and their distillation version Real-ISR methods, demonstrating quality improvement and the effectiveness of proposed approach.
\end{itemize}

\section{Related Work}
\label{sec:related}

\subsection{Real-World Image Super-Resolution}
To address the challenge of recovering real-world low-resolution (LR) observations with unknown degradations, the Real-ISR task has been introduced. Due to the complex degradations involved, Real-ISR has been a challenging problem for some time~\cite{ignatov2017dslr, liu2022blind, ji2020real}. Initial methods trained their models with simple downsampling techniques~\cite{kim2016accurate, zhang2018image}, which led to poor performance on real-world datasets. BSRGAN~\cite{zhang2021designing} and Real-ESRGAN~\cite{wang2021real} were among the first to introduce a more realistic synthesis pipeline for LR images, enabling deep learning methods to be applied to real-world scenarios. However, these GAN-based methods often suffer from artifacts and training instability.
With the emergence of the powerful pre-trained text-to-image generation model Stable Diffusion (SD)~\cite{rombach2022high}, many efforts have been made to leverage its strong generative capabilities for solving the Real-ISR problem, such as DiffBIR~\cite{lin2023diffbir}, StableSR~\cite{wang2024exploiting} and SeeSR~\cite{wu2024seesr}, and FaithDiff~\cite{chen2025faithdiff}. These SD-based methods improve fidelity and perceptual quality, but their application is limited due to the significant computational resources and time required. This is primarily due to the dozens to hundreds of timesteps involved in the diffusion denoising process.
\subsection{One-step Diffusion Models for Real-ISR}
To reduce the computational cost during inference, one-step diffusion methods have been proposed. These methods utilize model distillation techniques specifically designed for diffusion models, including progressive distillation~\cite{meng2023distillation, salimans2022progressive}, consistency models~\cite{song2023consistency}, distribution matching distillation~\cite{yin2024one, yin2024improved}, and variational score distillation (VSD)~\cite{wang2024prolificdreamer, nguyen2024swiftbrush}. In the context of Real-ISR, OSEDiff~\cite{wu2024one} employs the VSD strategy to achieve one-step diffusion based on a pre-trained SD model. Similarly, S3Diff~\cite{zhang2024degradation-s3diff} directly employs a distilled Stable Diffusion Turbo (SD-T) model for Real-ISR. Recent works including ADCSR~\cite{chen2025adversarial} and TSD-SR~\cite{dong2025tsd} also improve OSD performance and efficiency. InvSR\cite{yue2025arbitrary} offers a flexible sampling mechanism with arbitrary-steps (1-5) diffusion.

However, while time efficiency is improved, another dilemma arises: one-step diffusion methods struggle to generate diverse outputs balancing fidelity and realism—a critical requirement for real-world applications—unlike multi-step approaches that achieve this through progressive noise scheduling. In other words, current one-step diffusion methods for Real-ISR cannot control fidelity and realism as effectively as their multi-step counterparts. Some techniques, such as ControlNet models~\cite{zhang2023adding}, can enhance HR image generation by controlling semantic content but cannot achieve pixel-level control and require additional conditions, such as extra images or depth maps. Therefore, these techniques cannot be directly employed in one-step diffusion methods for Real-ISR. {OFTSR~\cite{zhu2024oftsr} achieves fidelity trade-offs through trajectory alignment distillation, but lacks degradation-aware mechanisms for real-world scenarios. PiSA-SR~\cite{sun2025pixel} controls fidelity and realism by training and inference with two different LoRA~\cite{hu2022lora} modules and two-step diffusion, which obviously increases the computational cost compared to using a single step.} Thus, a simple, efficient, and generalizable method for fidelity-realism control in OSD for Real-ISR remains essential.

\section{Methodology}
\label{sec:method}
In this section, we first reveal the characteristic of denoising network in one-step diffusion for image super-resolution, and then propose our algorithm.
\subsection{Preliminary: The Character of Denoising Network in One-step Diffusion for Real-ISR}
In SD, the latent diffusion process starts with encoding an image into a latent representation $z_0$ using a VAE encoder. The forward diffusion process then adds Gaussian noise to $z_0$ over $T$ steps via a Markov chain defined as:
\begin{align}
    q(\boldsymbol{z}_t|\boldsymbol{z}_{t-1}) = \mathcal{N}(\boldsymbol{z}_t;\sqrt{1-\beta_t}\boldsymbol{z}_{t-1},\beta_t \mathcal{I}) 
\end{align}
following a variance schedule $\beta_1, \ldots, \beta_T$. Here $\boldsymbol{z}_0 \sim q(\boldsymbol{z}_0)$. 
The forward process is: $\boldsymbol{z}_t = \alpha_t \boldsymbol{z}_0 + \sigma_t \boldsymbol{\epsilon}$, where $\alpha_t = \prod_{s=1}^t \sqrt{1 - \beta_s}$, $\sigma_t = \sqrt{1 - \alpha_t^2}$, and $\boldsymbol{\epsilon} \sim \mathcal{N}(\mathbf{0}, \mathcal{I})$. 
Here, the mean of $\boldsymbol{z}_t$ becomes $\alpha_t \boldsymbol{z}_0$, and the variance is $\sigma_t^2 \mathcal{I}$. With larger $t$, more noise is added, resulting in a greater deviation of the mean (scaled by $\alpha_t < 1$) and an increased variance ($\sigma_t^2$), distancing $\boldsymbol{z}_t$ from the original distribution $\boldsymbol{z}_0$.

The reverse process denoises $\boldsymbol{z}_t$ to recover $\boldsymbol{z}_{t-1}$:
\begin{align}
    p_\theta(\boldsymbol{z}_{t-1}|\boldsymbol{z}_{t}) = \mathcal{N}(\boldsymbol{z}_{t-1};\boldsymbol{\mu}_\theta(\boldsymbol{z}_{t}, t), \boldsymbol{\Sigma}_\theta(\boldsymbol{z}_{t}, t)), 
\end{align}
where a time-conditional U-Net $\epsilon_\theta$ predicts the noise $\boldsymbol{\epsilon}$ to estimate $\boldsymbol{\mu}_\theta$ and $\boldsymbol{\Sigma}_\theta$. Multi-step diffusion models iteratively refine $\boldsymbol{z}_T$ back to $\boldsymbol{z}_0$ over $T$ steps, while one-step diffusion methods, often distilled from multi-step models, predict the clean data directly from $\boldsymbol{z}_T$ in a single step, significantly reducing computation.

For Real-ISR tasks using SD-based OSD methods like~\cite{wu2024one}, the process from LR latent features $\boldsymbol{z}_L$ to HR latent features $\hat{\boldsymbol{z}}_H$ is formulated as a one-step denoising process:
\begin{align}
    \hat{\boldsymbol{z}}_H = F_{\theta}(\boldsymbol{z}_L; T, c_y) \triangleq \frac{\boldsymbol{z}_L - \beta_T \boldsymbol{\epsilon}_{\theta}(\boldsymbol{z}_L; T, c_y)}{\alpha_T},
    \label{equ:latent_mapping_function}
\end{align}
where $\boldsymbol{z}_L$ is the LR latent representation, $c_y$ is the text embedding, and $\epsilon_{\theta}$ is the denoising network predicting noise at timestep $T$. As shown in Fig.~\ref{fig:main_principle}(a), vanilla OSD methods learn a direct {latent feature} mapping from LR to HR images {with or without adding additional noise}.

In Real-ISR, images exhibit diverse degradation types and levels. SD-based methods use powerful generative image priors to recover LR images. However, OSD typically trains on all data with a fixed timestep $T$ with a constant noise level. This results in a model that generates a uniform amount of detail {and converges to a confined domain}, which may not suit images with varying degradation severity. 
{Tab.~\ref{tab:two_mini_table} (a) demonstrates the results of training OSEDiff on two different degradation pipelines (DP). `Orig.+ New Deg.' denotes a DP applying more degradations than the standard DP (`Orig.'). It indicates that higher degradation in training results in a greater emphasis on realism. This exposes OSD's flaw: by locking training to a single fixed T,
} the model optimizes for an \textbf{``average'' degradation}, yielding a limited generation flexibility that struggles to adapt its output to meet the specific scenario requirements.

In contrast, multi-step diffusion models provide greater flexibility in SR tasks. By selecting different timesteps $t$ during the inference stage, these models control the degree of noise adding and removing in diffusion, balancing fidelity and realism effectively.

Therefore, to overcome the inherent limitation of OSD, which is incapable of adjusting its generation levels to adapt to varying scenarios, we propose a novel training strategy for real-world SR applications that can flexibly control the generation realism during the inference stage.

\begin{table}[htbp]
\centering
\begin{minipage}{0.53\linewidth}
\centering
\captionsetup{type=table}

\resizebox{\linewidth}{!}{
\begin{tabular}{c|cc}
\toprule
Metrics & Orig. & Orig.+New Deg.  \\
\midrule
PSNR $\uparrow$ & \textbf{25.15} & 24.59  \\
MANIQA $\uparrow$ & 0.6326 & \textbf{0.6462}  \\
\bottomrule
\end{tabular}}
\label{tab:deg_gen}
\caption*{(a)}
\end{minipage}
\hfill
\begin{minipage}{0.43\linewidth}
\centering
\captionsetup{type=table}
\resizebox{\linewidth}{!}{
\begin{tabular}{c|ccc}
\toprule
$M_L$ & L1 & MSE & \textbf{Cosine Sim.}  \\
\midrule
SSIM $\uparrow$ & 0.60 & 0.54 & {\textbf{0.78}}  \\
DISTS $\uparrow$ & 0.15 & 0.11 & \textbf{0.42}  \\
CLIPIQA $\uparrow$ & 0.06 & 0.05 & \textbf{0.27}  \\
\bottomrule
\end{tabular}}
\caption*{(b)}
\end{minipage}
\captionsetup{justification=centering}
\caption{{(a) Influence of degradation degree in training data. (b) $|\text{Spearman coefficient}|\uparrow$ comparison of different $M_L$ distances and image quality metrics.}}
\label{tab:two_mini_table}
\end{table}

\begin{figure}[htbp]
\scriptsize
\centering
    \begin{minipage}{0.39\linewidth}
        \centering
        \includegraphics[width=\linewidth]{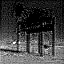}
        {Noisy Latent}
    \end{minipage}
    \begin{minipage}{0.58\linewidth}
        \centering
           \setlength{\tabcolsep}{1pt}
        \begin{tabular}{ccc}
            \includegraphics[width=0.32\linewidth]{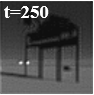} & 
            \includegraphics[width=0.32\linewidth]{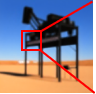} &
            \includegraphics[width=0.32\linewidth]{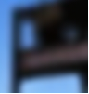}\\
            \includegraphics[width=0.32\linewidth]{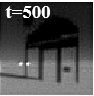} & 
            \includegraphics[width=0.32\linewidth]{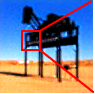} &
            \includegraphics[width=0.32\linewidth]{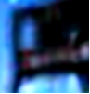}\\
            {Denoised Latent} & {VAE Decoded}& {Zoomed} \\
        \end{tabular}
    \end{minipage}
\caption{{Influence of different timesteps $t$ using SD-turbo.}}
\label{fig:noise_t}
\end{figure}

\subsection{Latent Domain Grouping}\label{sec:LDG}
{To achieve dynamic fidelity-realism trade-offs control in OSD generated SR results,} we focus on the most basic condition in denoising network, \ie, timestep condition.
Unlike prompts from a text encoder or a vision encoder, the timestep condition is an unremovable component in the diffusion process. At the same time, it controls the mean and variance of noisy latent feature $z_{T}$. With the larger mean and variance difference between $z_{T}$ and $z_{0}$, more contents will be generated during the denoising process. {Fig.~\ref{fig:noise_t} shows an example of influence of different $t$. In the foundational model SD-turbo, the higher timestep value during the diffusion process usually reflects the the higher capability generating.}

Therefore, to easily control the generation degree, we propose a latent domain grouping (LDG) strategy. Recall Eq.~\eqref{equ:latent_mapping_function}, we do not use a single fixed timestep $T$, but choose a timestep $t$ according to a metrics:
\begin{align}
    \hat{\boldsymbol{z}}_H &= F_{\theta}(\boldsymbol{z}_L; t, c_y),
    \label{equ:latent_mapping_function-2} \\
t &=  k \cdot (n - \left\lfloor \frac{n \cdot (M_L - M_{L\text{-min}})}{(M_{L\text{-max}} - M_{L\text{-min}}) } \right\rfloor), ~k\in\mathbb{Z}^+,
\label{equ:grouping}
\end{align}
where $M_{L}$ denotes a latent metric that can perceive the ``level of degradation'' of features in latent domain, $M_{L\text{-min}}$ is the minimum value of $M_{L}$ in training data, $k$ is interval of timestep, $\lfloor.\rfloor$ denotes the maximum integer no larger than the entry inside,   $n$ is number of groups for timestep.

To employ this strategy both on SD and distillation version of SD, \ie, SD-Turbo (SDT), which distilled a four specific steps from the original 1000-step diffusion process, we set $n$ to be $\leq4$ and $k=250$.

By the grouping strategy, denoising network can learn different degrees of generation according to timestep. {In the training stage, grouping is based on the  $M_{L}$ described in the next subsection.}
In the inference stage, we can easily choose a timestep to control the level of realism for SR in different scenarios. Furthermore, due to our grouping strategy, the realism level increases monotonically with the timestep.

\subsection{Latent Metric for Denoising Network}\label{sec:Latent Metric}

The latent metric $M_{L}$ is designed to enable the denoising network in the latent domain to perceive the ``level of degradation'' of the low-resolution features $\mathbf{z}_L$. Here, we define the ``level of degradation'' as the extent to which $\mathbf{z}_L$ deviates from its HR counterpart $\mathbf{z}_H$. Thus, the definition of the latent metric $M_{L}$ should reflect the characteristics of $\mathbf{z}_L$ that indicate this degradation.

\begin{figure}
    \centering
    \includegraphics[width=0.9\linewidth, trim=0 0 0  0, clip]{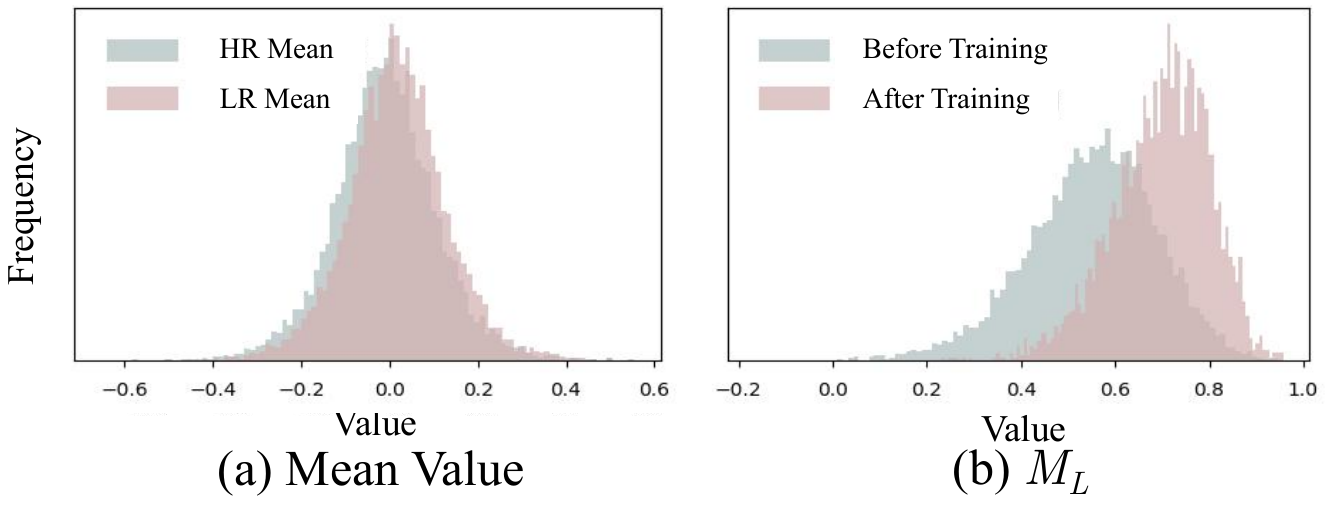}
    \caption{Distribution of (a) mean value of LR and HR training images in the latent domain, (b) the $M_L$ metric in the latent domain of VAE before and after training.}
    \label{fig:three_hist}
\end{figure}

A simple choice might be to use the difference between the mean values of $\mathbf{z}_L$ and $\mathbf{z}_H$, as the forward diffusion process scales the mean of $\mathbf{z}_L$ over time. However, as shown in Fig.~\ref{fig:three_hist}(a), the mean values of $\mathbf{z}_L$ and $\mathbf{z}_H$ are similar across the training data. This suggests that mean difference fails to effectively capture the degradation level.

Instead, we use cosine similarity (CS) as $M_{L}$:
\begin{align}
M_L = \frac{\mathbf{z}_L \cdot \mathbf{z}_H}{\|\mathbf{z}_L\| \|\mathbf{z}_H\|}.
\label{equ:cs_score}
\end{align}
Cosine similarity is a widely adopted measure in representation learning (e.g., contrastive learning~\cite{chen2020simple}). It can quantify the divergence between high-dimensional latent features $\mathbf{z}_L$ and $\mathbf{z}_H$, which can reflect high-level changes caused by degradation. 
As shown in Fig.~\ref{fig:three_hist}(b), most cosine similarity values lie between 0 and 1, providing a clear range to distinguish varying degrees of degradation for the denoising network.

{Different distances inherently introduce a certain preference or bias, to evaluate the bias of different $M_{L}$, we calculate correlation coefficient ($|\text{Spearman Corr.}| \uparrow$) of CS, L1, and MSE with different metrics in Table~\ref{tab:two_mini_table} (b). The metrics are calculated between LR and HR images. The correlation reflects the degree of association between various $M_{L}$ and different metrics. CS exhibits a higher correlation coefficient with objective (SSIM), perceptual (DISTS), and semantic (CLIPIQA) metrics compared to other distances in latent space. \textbf{More visualization results can be seen in Supplementary Materials (SM).}}

\subsection{Metric Estimation}
Since $\mathbf{z}_H$ is introduced as a latent metric, we can only calculate it during training. In the inference stage, beyond manually choosing a timestep, we also consider an adaptive timestep selection, i.e., estimating an $M_{L}$ for each LR image. Benefiting from the powerful representation ability of the pre-trained model, we use features from intermediate layers as input and a simple MLP as a metric estimation module (MEM), operating independently of OSD training.

\subsection{Degradation-aware Sampling Distillation}
 {Previous VSD method for SR~\cite{wu2024one}, utilizes the regularization network (a pre-trained SD) that sampled timesteps across a wide range (20-980). This aims to generate regularization latent features and, in turn, optimize the distribution of the OSD network. However, to better integrate the distillation process with the concept of degradation in the latent space, we propose a Degradation-Aware Sampling (DAS) strategy. DAS redefines how timesteps are sampled in the pre-trained model, adaptively aligning this process with our LDG framework to provide explicit control over regularization strength. The DAS can be written as:
\begin{align}
    t_r &=S(\max(20, t - k), \min(980, t + k))
    \label{equ:vsd_sample_time} 
\end{align}
where $t_r$ is the sample timestep for regularization network, $t$ is the chosen timestep in OSD network by LDG, $S(t_{min},t_{max})$ denotes uniformly random sampling of an integer from the range $[t_{min},t_{max}]$.

  By applying DAS, the degradation grouping information is delivered from LDG, thereby aligning this process with LDG and control over regularization strength.}
\subsection{Visual Prompt Injection Module}

In previous SD-based super-resolution (SR) methods like OSEDiff~\cite{wu2024one} and SeeSR~\cite{wu2024seesr}, text encoders, sometimes paired with a vision-language model (VLM) as a text prompt extractor, improve non-reference (NR) metrics, which assess realism, but often constrain full-reference (FR) metrics, which assess fidelity to the ground truth. This trade-off arises because text prompts provide high-level semantic guidance to enhance realism, yet they may compromise structural accuracy.

LR feature $\mathbf{z}_L$ solely from the VAE encoder offers limited semantic information. Without additional context, the conditioned U-Net struggles to generate high-quality outputs. Text prompts attempt to bridge this gap by injecting external semantic cues, but they come with drawbacks: VLMs increase computational costs, and the prompts may not fully align with the image’s content. Some methods like S3Diff~\cite{zhang2024degradation-s3diff} use fixed text to reduce complexity, yet still struggle to balance NR and FR metrics.

To address these issues, we propose the Visual Prompt Injection Module (VPIM), which replaces conventional text prompts with degradation-aware visual tokens. VPIM substitutes the text encoder (typically a CLIP text model) with a CLIP vision model and an MLP layer for dimension alignment. The LR image serves as its input, \ie,  visual prompt, and the output is fed into the cross-attention of the U-Net. By adopting VPIM, we eliminate the need for VLMs, reduce computational overhead, and provide the U-Net with image-specific semantic information directly from the LR input. The visual prompt is tied to the image’s pixel characteristics, leading to improvement in both fidelity and realism.

With the combination of LDG, latent metric, DAS, and VPIM, we proposed a Realism control one-step diffusion (RCOD) framework, a realism-flexible one-step diffusion model with enhanced performance that can be applied to various recent mainstream one-step diffusion Real-ISR methods, thereby improving their capabilities. \textbf{We provide a pseudo-code example of our RCOD in SM.}

\begin{table*}[t]
  \centering
    \setlength{\tabcolsep}{2pt} 
  \resizebox{1\textwidth}{!}
  {
  \begin{tabular}{c|c|ccccccccc}
    \toprule
    Datasets & Methods &  PSNR↑  &  SSIM↑  &  LPIPS↓  &  DISTS↓  &  NIQE↓  &  MUSIQ↑  &  MANIQA↑  &  CLIPIQA↑  & FID↓ \\
    \midrule
    \multirow{13}[2]{*}{DrealSR} & SinSR & \underline{28.36} & 0.7515 & 0.3665 & 0.2485 & 6.991 & 55.33 & 0.4884 & 0.6383 & 170.57 \\
          & PiSA-SR & 28.31 & 0.7804 & 0.2960 & 0.2169 & 6.200 & 66.11 & 0.6156 & 0.6970 & 130.61 \\
          & TSD-SR & 25.67 & 0.7132 & 0.3538 & 0.2459 & 5.991 & 65.99 & 0.6327 & 0.7137 & 171.36 \\
          & InvSR & 28.33 & 0.7502 & 0.3678 & 0.2481 & 6.941 & 55.27 & 0.4900 & 0.6385 & 170.57 \\
          & OSEDiff & 27.92 & \underline{0.7835} & \underline{0.2968} & 0.2165 & 6.490 & 64.65 & 0.5899 & 0.6963 & 135.30 \\
          & RCOD$_\text{O}$-Fid. & \textbf{28.90} & \textbf{0.7906} & \textbf{0.2919} & 0.2186 & 6.817 & 66.72 & 0.6275 & 0.7023 & 131.52 \\
          & RCOD$_\text{O}$-Neu. & 28.30 & 0.7775 & 0.3080 & 0.2306 & 6.469 & \underline{68.03} & \textbf{0.6385} & 0.7179 & 139.92 \\
          & RCOD$_\text{O}$-Real. & 27.59 & 0.7600 & 0.3389 & 0.2499 & 6.172 & \textbf{68.19} & 0.6295 & 0.7325 & 158.16 \\
          & S3Diff & 27.54 & 0.7491 & 0.3109 & \textbf{0.2099} & 6.212 & 63.94 & 0.6134 & 0.7130 & \textbf{118.64} \\
          & RCOD$_\text{S}$-Fid. & 28.09 & 0.7800 & 0.3002 & \underline{0.2149} & 5.871 & 65.74 & 0.6174 & 0.6963 & 136.62 \\
          & RCOD$_\text{S}$-Neu. & 27.48 & 0.7526 & 0.3256 & 0.2251 & \underline{5.636} & 67.40 & \underline{0.6339} & \underline{0.7278} & 138.87 \\
          & RCOD$_\text{S}$-Real. & 26.95 & 0.7190 & 0.3607 & 0.2414 & \textbf{5.448} & 67.58 & 0.6317 & \textbf{0.7478} & 145.36 \\
          & RCOD$_\text{S}$-Adap. & 27.83 & 0.7661 & 0.3098 & 0.2181 & 5.768 & 66.32 & 0.6223 & 0.7110 & \underline{135.61} \\
    \midrule
    \multirow{13}[2]{*}{RealSR} & SinSR & \textbf{26.28} & 0.7347 & 0.3188 & 0.2353 & 6.287 & 60.80 & 0.5385 & 0.6122 & 135.93 \\
          & PiSA-SR & 25.50 & 0.7417 & \underline{0.2672} & 0.2044 & 5.500 & 70.15 & 0.6560 & 0.6702 & 124.09 \\
          & TSD-SR & 23.41 & 0.6886 & 0.2805 & 0.2183 & 5.093 & \textbf{70.77} & 0.6311 & \textbf{0.7193} & 114.56 \\
          & InvSR & 24.13 & 0.7125 & 0.2871 & 0.2123 & 5.626 & 68.54 & 0.6619 & 0.6790 & 138.88 \\
          & OSEDiff & 25.15 & 0.7341 & 0.2921 & 0.2128 & 5.648 & 69.09 & 0.6326 & 0.6693 & 123.49 \\
          & RCOD$_\text{O}$-Fid. & \underline{26.01} & \textbf{0.7427} & 0.2796 & 0.2103 & 5.911 & 70.25 & 0.6647 & 0.6866 & 121.40 \\
          & RCOD$_\text{O}$-Neu. & 25.39 & 0.7264 & 0.2939 & 0.2190 & 5.497 & 70.34 & \underline{0.6750} & 0.7022 & 127.74 \\
          & RCOD$_\text{O}$-Real. & 24.62 & 0.7011 & 0.3296 & 0.2375 & 5.341 & \underline{70.76} & 0.6650 & 0.7084 & 143.74 \\
          & S3Diff & 25.18 & 0.7269 & 0.2721 & \underline{0.2005} & 5.269 & 67.82 & 0.6437 & 0.6727 & \textbf{105.14} \\
          & RCOD$_\text{S}$-Fid. & 25.42 & \underline{0.7392} & \textbf{0.2647} & \textbf{0.1976} & 5.095 & 69.46 & 0.6605 & 0.6509 & 113.93 \\
          & RCOD$_\text{S}$-Neu. & 24.78 & 0.7130 & 0.2855 & 0.2073 & \underline{5.024} & 70.55 & \textbf{0.6757} & 0.6886 & 114.80 \\
          & RCOD$_\text{S}$-Real. & 24.08 & 0.6759 & 0.3228 & 0.2236 & \textbf{4.900} & 70.65 & 0.6719 & \underline{0.7086} & 120.00 \\
          & RCOD$_\text{S}$-Adap. & 25.23 & 0.7313 & 0.2714 & 0.2010 & 5.033 & 69.72 & 0.6646 & 0.6622 & \underline{112.93} \\
    \bottomrule
    \bottomrule
  \end{tabular}%
  }
  \caption{Quantitative comparison with state-of-the-art methods on both synthetic and real-world benchmarks. The best and second best results are highlighted in \textbf{bold} and \underline{underline}, respectively.}
  \label{tab:main_3data}%
\end{table*}

\section{Experiments}
\label{sec:exp}
\subsection{Experiments Setups} 
\noindent
\textbf{Datasets:} Following the training and testing settings of prior works\cite{wu2024seesr,wu2024one,zhang2024degradation-s3diff}, we employ LSDIR \cite{li2023lsdir} and the first 10K face images from FFHQ \cite{ffhq} for training and degradation pipeline of Real-ESRGAN \cite{wang2021real} for LR image synthesizing.
The synthetic test data use cropped $512\times512$ synthetic data from DIV2K-Val \cite{div2k} and degraded using the Real-ESRGAN pipeline \cite{wang2021real}. The real-world data include LR-HR pairs from RealSR \cite{realsr} and DRealSR \cite{drealsr}, both with sizes of $128\times128$-$512\times512$ for LR-HR pairs.

\noindent
\textbf{Evaluation Metrics:} We employ widely used FR and NR metrics. FR metrics include PSNR, SSIM \cite{ssim}, LPIPS \cite{lpips}, and DISTS \cite{dists}. NR metrics include NIQE \cite{niqe}, MANIQA-pipal \cite{maniqa}, MUSIQ \cite{musiq}, and CLIPIQA \cite{clipiqa}.

\noindent
\textbf{Method Comparison:} 
We compare our method with state-of-the-art methods (SOTA) in three categories: multi-step diffusion Real-ISR methods, including StableSR~\cite{wang2024exploiting}, ResShift~\cite{yue2023ResShift}, DiffBIR~\cite{lin2023diffbir}, and SeeSR~\cite{wu2024seesr}; one-step diffusion Real-ISR methods, such as SinSR~\cite{wang2023sinsr}, OSEDiff~\cite{wu2024one}, S3Diff~\cite{zhang2024degradation-s3diff}, TSD-SR~\cite{dong2025tsd}, PiSA-SR (default version)~\cite{sun2025pixel}, and InvSR~\cite{yue2025arbitrary}; and GAN-based methods, including BSRGAN~\cite{zhang2021designing} and Real-ESRGAN~\cite{wang2021real}. We quantitatively compare recent SOTA OSD methods in Table \ref{tab:main_3data} on real-world data. \textbf{More qualitative comparisons and the full table including synthetic data, multi-step diffusion, and GAN-based methods are in SM.}

\subsection{Implement Details} 
\noindent
\textbf{Model Setting:} 
To verify our framework, we select two types of recent SOTA OSD Real-ISR methods: OSEDiff~\cite{wu2024one}, which is distilled from a pre-trained multi-step Stable Diffusion (SD) model, and S3Diff~\cite{zhang2024degradation-s3diff}, which directly uses SD-T (a distilled version of SD), as our base models. When RCOD is applied to them, they are named RCOD$_\text{O}$ and RCOD$_\text{S}$, respectively. The choice of $n=3$ corresponds to three distinct generation levels in the inference stage: Fidelity, Neutral, and Realism. \ie, $t=250,~500, \text{and~}750$ during inference. {Since the S3Diff is not directly distilled from a multi-step diffusion, we do not apply DAS on it.} RCOD$_\text{S}$-Adap. employs MEM during inference. The input to the MEM in this case is $\boldsymbol{z}_L$ and the features in the last layer degradation estimation model used in~\cite{zhang2024degradation-s3diff}. The MEM is trained after RCOD$_\text{S}$ training, utilizing the same training data. \textbf{More details please refer to SM.}

\begin{figure*} 
  \centering
    \resizebox{1\textwidth}{!}{%
    \begin{tabular}{cc}
\hspace{-0.4cm}
\begin{adjustbox}{valign=t}
\begin{tabular}{c}
\includegraphics[width=0.413\textwidth]{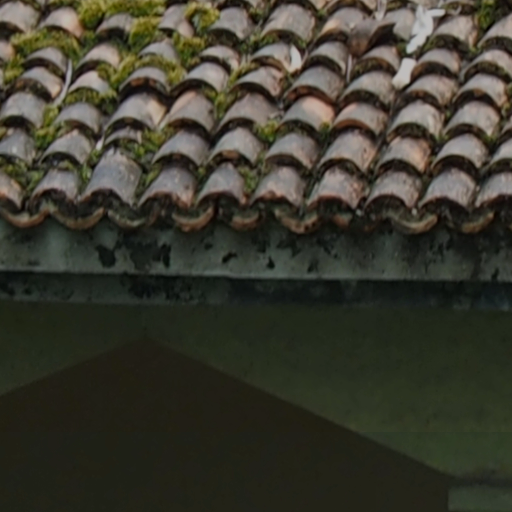} \\
P1140134 HR
\end{tabular}
\end{adjustbox}
\hspace{-0.46cm}

\begin{adjustbox}{valign=t}
\begin{tabular}{cccc}
\includegraphics[width=0.135\textwidth]{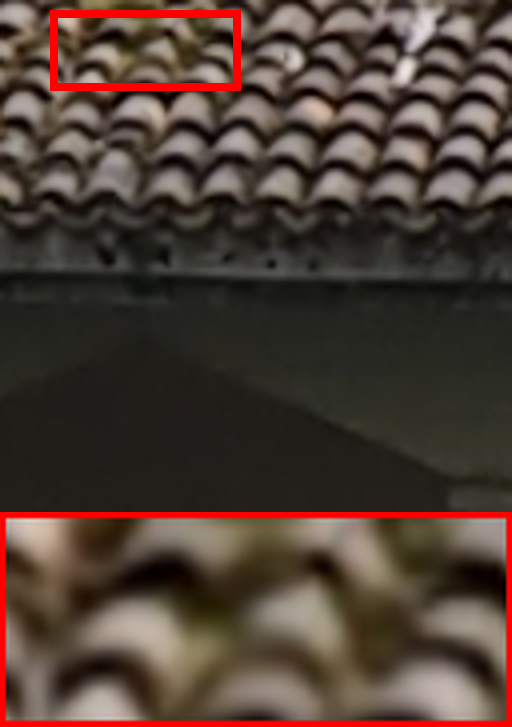} \hspace{-4mm} &
\includegraphics[width=0.135\textwidth]{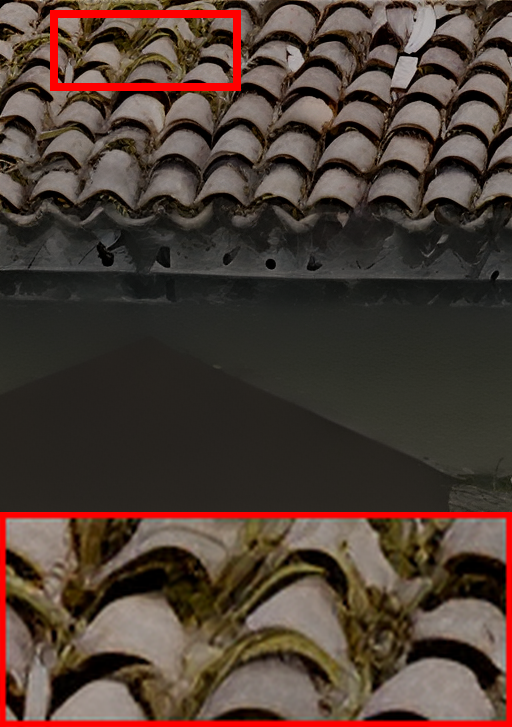} \hspace{-4mm} &
\includegraphics[width=0.135\textwidth]{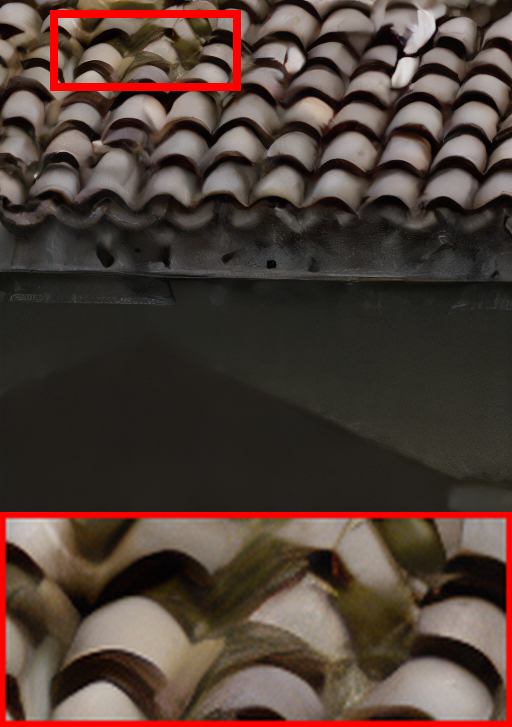} \hspace{-4mm} &
\includegraphics[width=0.135\textwidth]{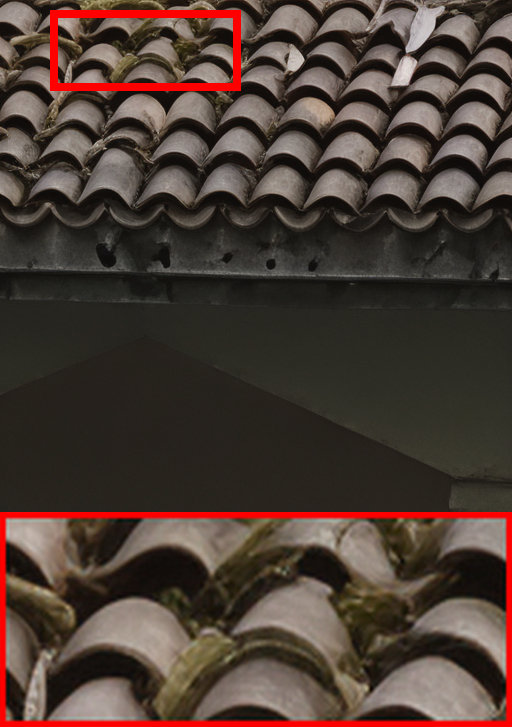} \hspace{-4mm} \\
LR \hspace{-4mm} &
StableSR \hspace{-4mm} &
SinSR \hspace{-4mm} &
PiSA-SR\hspace{-4mm} \\
\includegraphics[width=0.135\textwidth]{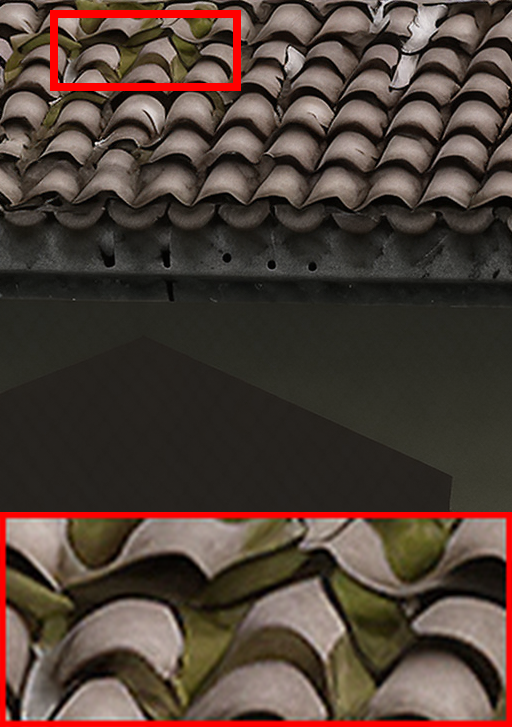} \hspace{-4mm} &
\includegraphics[width=0.135\textwidth]{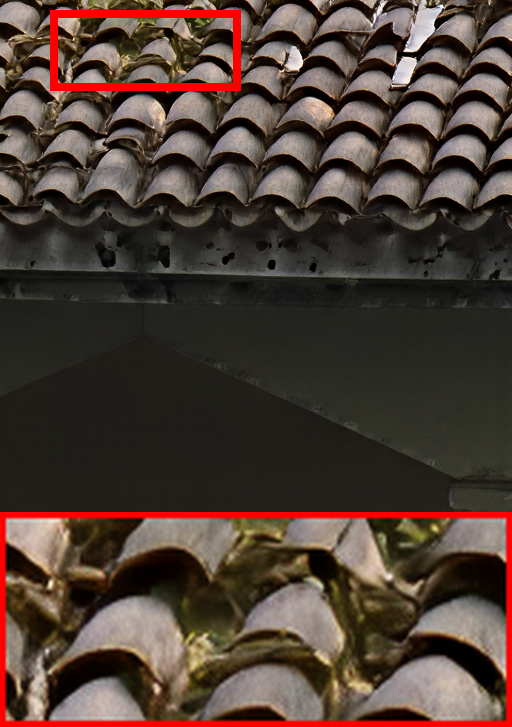} \hspace{-4mm} &
\includegraphics[width=0.135\textwidth]{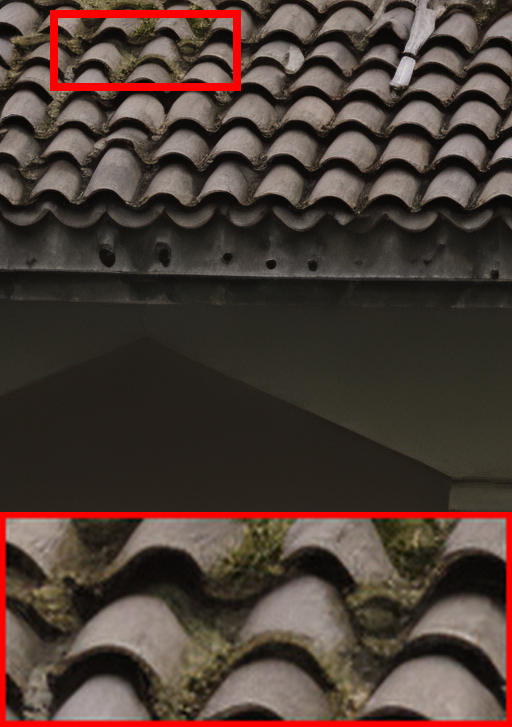} \hspace{-4mm} &
\includegraphics[width=0.135\textwidth]{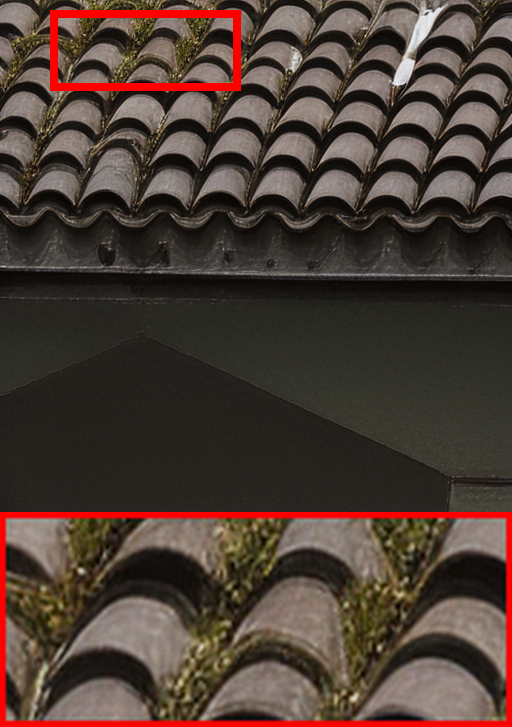} 
\hspace{-4mm} \\
InvSR \hspace{-4mm} &
TSD-SR\hspace{-4mm} &
OSEDiff \hspace{-4mm} &
RCOD$_\text{O}$-Real. \hspace{-4mm} \\ 
\end{tabular}
\end{adjustbox}
\end{tabular}
}
\caption{Visual comparison ($\times$4) of RCOD$_\text{O}$-Real. with other methods on DRealSR data.}
\label{fig:main_vis_compare}
\end{figure*}

\subsection{Comparison with State-of-the-Arts}
\noindent
\textbf{Quantitative Comparisons:} We can observe in Table \ref{tab:main_3data}: 
$i$) By applying RCOD, on each dataset, the ``-Fid.'' versions ($t=250$) have better full-reference (FR) metrics such as PSNR, SSIM, LPIPS, and FID, while keeping the no-reference (NR) metrics relatively ordinary. In contrast, the ``-Real.'' versions ($t=750$) achieve obviously higher NR metrics like MANIQA, MUSIQ, and CLIPIQA. Most ``-Neu.'' versions ($t=500$) fall within the middle range of the previous two versions. This illustrates that we can effectively and simply control the realism level (usually measured by perceptual NR metrics) during the inference stage, and that the realism level increases monotonically with the timestep.
$ii$) RCOD$_\text{S}$-Adap. has relatively balanced metrics between RCOD$_\text{S}$-Fid. and RCOD$_\text{S}$-Neu.. This indicates that most estimated $M_L$ values are closer to 1 than to 0. This roughly matches the cosine similarity value distributions in Fig.~\ref{fig:three_hist} (b).
$iii$) When applying RCOD, RCOD$_\text{O}$-Fid. and RCOD$_\text{S}$-Adap. perform better than their original methods (OSEDiff and S3Diff, respectively) on most metrics, including PSNR, SSIM, LPIPS, MANIQA, and MUSIQ. Even with a preference for fidelity in FR metrics, RCOD$_\text{O}$-Fid. shows superior performance on some NR metrics, such as MANIQA and MUSIQ, compared to the original OSEDiff method on real-world data. 
Additionally, RCOD$_\text{S}$-Real. usually achieves the best NR metrics (NIQE and CLIPIQA).
$iv$) S3Diff shows better performance on the perceptual quality metric DISTS. This may arise from the negative online prompting (NOP) used in training, which provides a more accurate concept of high quality. However, since the text encoder and text prompt are replaced by VPIM in our RCOD$_\text{S}$, the NOP is also removed. Despite this, our RCOD$_\text{S}$-Adap. performs better on other NR metrics.

\noindent
\textbf{Time Efficiency:} In Table \ref{tab:abla_time}, we compare inference time and trainable parameters. All methods are tested on an A100 GPU with a $512\times512$ input image. RCOD keeps similar time efficiency and trainable parameters as the original methods while have higher PSNR and MANIQA. RCOD$_\text{O}$ even inferences faster as the text extractor is removed. 

\begin{figure}
\scriptsize
\centering
\setlength{\tabcolsep}{1pt} %
\begin{tabular}{c}
\includegraphics[width=1\linewidth, trim=0 62 0 214, clip]{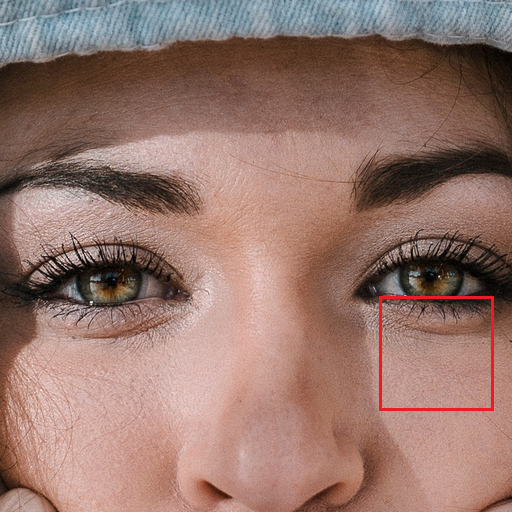} \\
DIV2K-Val-0855\_pch\_00018
\end{tabular}
\begin{tabular}{ccc}
\includegraphics[width=0.155\textwidth,trim=390 112 22 300, clip]{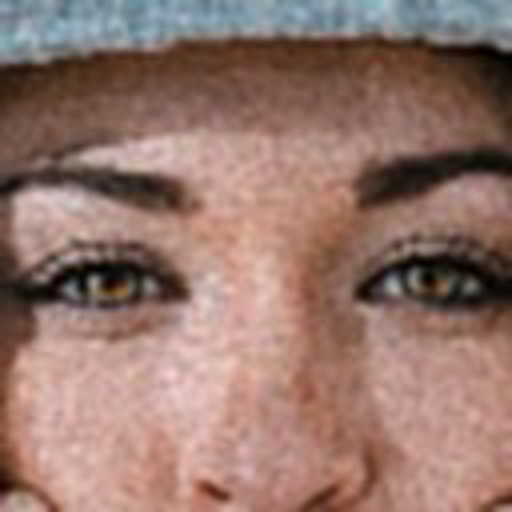} &
\includegraphics[width=0.155\textwidth,trim=390 112 22 300, clip]{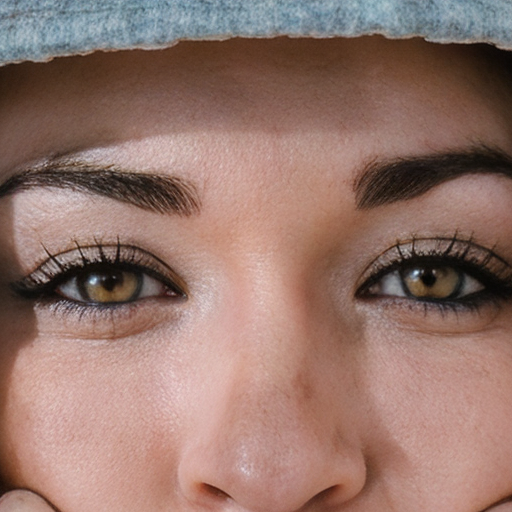} &
\includegraphics[width=0.155\textwidth,trim=390 112 22 300, clip]{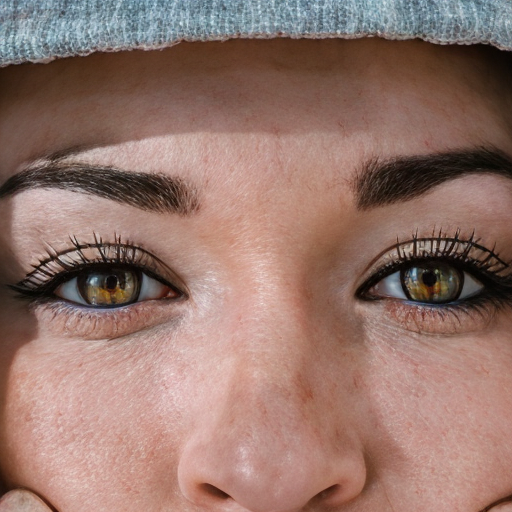} \\ 
LR ($M_L$=0.621) &
S3Diff &
RCOD$_\text{S}$-Adap. \\
SSIM / MUSIQ &
0.5088 / 76.78 &
0.5069 / 77.96 \\
\includegraphics[width=0.155\textwidth,trim=390 112 22 300, clip]{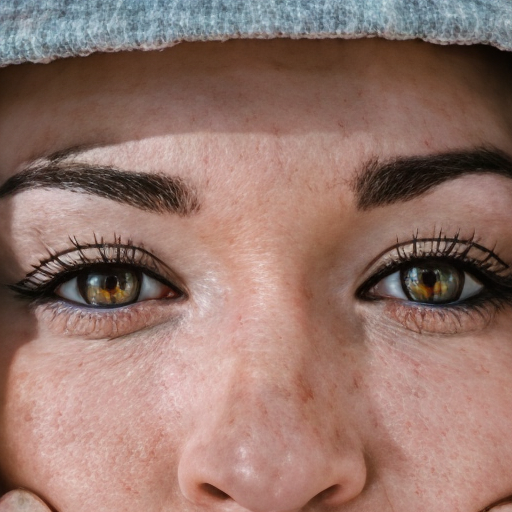} &
\includegraphics[width=0.155\textwidth,trim=390 112 22 300, clip]{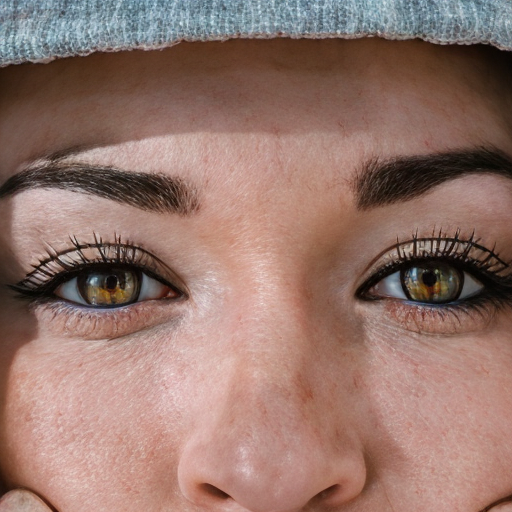} &
\includegraphics[width=0.155\textwidth,trim=390 112 22 300, clip]{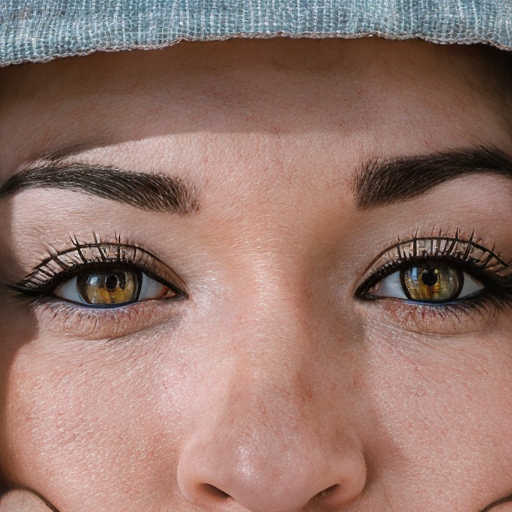} \\
RCOD$_\text{S}$-Fid. &
RCOD$_\text{S}$-Neu. &
RCOD$_\text{S}$-Real. \\
\textbf{0.5172} / 77.21 &
0.5069 / 77.96 &
0.5045 / \textbf{78.19} \\
\end{tabular}
\caption{Visual comparison ($\times$4) of realism controls during RCOD$_\text{S}$ inference stage. The best results are in \textbf{bold}.}
\label{fig:step_vis_compare}
\end{figure}

\noindent
\textbf{Qualitative Comparisons:}
Fig. \ref{fig:main_vis_compare} compares the visual qualities of different Real-ISR methods.  
RCOD$_\text{O}$-Real. demonstrates its ability to recover more detailed and natural textures. Fig.~\ref{fig:step_vis_compare} illustrates the changes in visual effects as $t$ increases, \ie, from RCOD$_\text{S}$-Fid. ($t=250$) to RCOD$_\text{S}$-Real. ($t=750$). As $t$ increases during the inference stage, more skin texture and wrinkles are recovered. RCOD$_\text{S}$-Adap. chooses a proper $t=500$ in this case, where the $M_L$ range is $[0.5, 0.75)$ according to Eq.~\ref{equ:grouping}. The $M_L$ of the LR image (0.621) falls within this range.

\noindent
\textbf{Ablation Study}
We performed a series of ablation studies of the RCOD framework, the details can be found in SM.

\section{Conclusion}
\label{sec:conclusion}
We propose RCOD, a framework that enhances one-step diffusion methods for Real-ISR through flexible realism control. RCOD employs latent grouping with degradation-aware sampling during distillation and introduces a robust latent metric enabling denoising networks to assess degradation levels. Applied to two distinct one-step diffusion methods, RCOD achieves superior super-resolution performance across most FR and NR metrics while maintaining computational efficiency. We believe RCOD holds promise for diverse Real-ISR scenarios with varying requirements.

\begin{table}[H]
  \centering
    \resizebox{1\linewidth}{!}{
    \begin{tabular}{c|ccccc}
    \toprule
         Metrics & PiSA-SR-adj. & OSEDiff & \textbf{RCOD$_\text{O}$-Fid.} & S3Diff & \textbf{RCOD$_\text{S}$-Adap.} \\
    \midrule
    PSNR $\uparrow$ & 28.31 & 27.92 & {\textbf{28.90}} & 27.54 & 27.83 \\
    MANIQA $\uparrow$ & 0.6156 & 0.5899 & \textbf{0.6275} & 0.6134 & {0.6223} \\
    \midrule
    Infer. Time (s) & 0.13  & 0.11  & {\textbf{0.09}} & 0.28  & 0.28 \\
    \#Trainable Param. (M) & \textbf{8.1} & {{8.5}} & 9.5 & 34.5  & 35.5 \\
    \bottomrule
    \end{tabular}%
 }
   \caption{Efficiency comparison on an NVIDIA A100 GPU. The best results are highlighted in bold.}
  \label{tab:abla_time}%
\end{table}%



\begin{appendices}
\input{SM}
\end{appendices}
{
    \bibliography{main}
}
\end{document}

%% file: SM.tex
\section{Supplementary Material for ``\textit{Realism Control One-step Diffusion for Real-World Image Super-Resolution}''}

\textbf{In the supplementary material, we provide the following contents:}
\begin{itemize}
    \item Algorithm Details: Pseudo-code of our RCOD$_\text{O}$ training strategy (referring to Section \textit{Methodology} in the main paper).
    \item Implement Details: Hyper-parameters setting and experimental environments (As a complement to Section \textit{Implement Details} in the main paper).
    \item Experiments: Full table of quantitative comparisons and more visual Comparisons (as a complement to Section \textit{Experiments} in the main paper).
    \item Ablation Study (referring to Section \textit{Ablation Study} in the main paper).
\end{itemize}

\subsection{Algorithm Details}


\noindent\textbf{Pseudo-code} The pseudo-code of our RC-OSD training strategy is summarized as Algorithm \ref{alg:OSEDiff_VPIM_MEM}. In inference stage, we do not need to compute $M_L$. Instead, we directly calculate $\hat{\boldsymbol{z}}_H = F_{\theta}(\boldsymbol{z}_L; t, c_y)$, where $t$ a user-specified parameter that controls the level of realism for SR output.
The choice of $t$ depends on the different application scenarios and is discussed in Subsection ``Latent Domain Grouping''.


\begin{algorithm*}
\footnotesize
\caption{{Training Scheme of RCOD$_\text{O}$. Differences from the original OSEDiff~\cite{wu2024one} are indicated by a \colorbox{lightgreen}{green}.}}
\label{alg:OSEDiff_VPIM_MEM}

\KwIn{Training dataset $\mathcal{S}$, pretrained SD parameterized by $\phi$, including VAE encoder $E_\phi$, latent diffusion network $\epsilon_{\theta}$, and VAE decoder $D_\phi$, VPIM $Y$, training iteration $N$}

Initialize generator $G_{\theta}$ with parameters: $E_{\theta} \gets E_\phi$ with trainable LoRA; $\epsilon_{\theta} \gets \epsilon_\phi$ with trainable LoRA; fixed parameters $D_{\theta} \gets D_\phi$; fixed parameters $Y_{\theta} \gets Y_\phi$; random initialize trainable parameters $\text{MLP}_{\theta}$ 
    

\For{$i \gets 1$ \KwTo $N$}{
    Sample low-resolution image $\boldsymbol{x}_L$ and corresponding high-resolution image $\boldsymbol{x}_H$ from $\mathcal{S}$

    \tcc{Encode LR and HR images}
    $(z_L, z_H) \gets E_{\theta}(\boldsymbol{x}_L, \boldsymbol{x}_H)$ 
    
    \tcc{Acquire degradation-aware visual tokens using VPIM}
    
    \colorbox{lightgreen}{$c_y \gets Y(\text{MLP}(z_L))$}

    \tcc{Latent domain grouping}
    \colorbox{lightgreen}{$M_L \gets \text{Equ. (6)}$; $t \gets \text{Equ. (5)}$}


    \tcc{Denoising and decoding}
    $\hat{z}_H \gets F_{\theta}(z_L; c_y, t)$;
    $\hat{\boldsymbol{x}}_H \gets D_{\theta}(\hat{z}_H)$

    \tcc{Data consistency loss}
    $\mathcal{L}_{\mathrm{data}} \gets \mathcal{L}_{\mathrm{MSE}}(\hat{\boldsymbol{x}}_H, \boldsymbol{x}_H) + \lambda_1 \mathcal{L}_{\mathrm{LPIPS}}(\hat{\boldsymbol{x}}_H, \boldsymbol{x}_H)$

    \tcc{{regularization with degradation-aware sampling}}
    
    Sample noise $\epsilon \sim \mathcal{N}(0, I)$ 
    
\colorbox{lightgreen}{Sample $t_r$ according to Equ. (7) }
    
    $\hat{z}_t \gets \alpha_{t_r} \hat{z}_H + \sigma_{t_r} \epsilon$; $z_{\phi} \gets \operatorname{stopgrad}(F_{\phi}(\hat{z}_t; c_y))$;
    $z_{\phi'} \gets \operatorname{stopgrad}(F_{\phi'}(\hat{z}_t; c_y))$;
    $\omega \gets 1 / \mathrm{mean}(\|z_{\phi} - \hat{z}_H\|)$;
    $\mathcal{L}_{\mathrm{reg}} \gets \omega \|z_{\phi'} - z_{\phi}\|$

    \tcc{Regularizer fine-tuning loss}
      Sample noise $\epsilon \sim \mathcal{N}(0, I)$ 
    
    Sample $t$ from \{1,...,T\} 
    $\boldsymbol{z}_t \gets \alpha_t \cdot \operatorname{stopgrad}(\hat{z}_H) + \sigma_t \epsilon$ 
    
    $\mathcal{L}_{\mathrm{diff}} \gets \mathcal{L}_{\mathrm{MSE}}(\epsilon_{\phi'}(z_t; t, c_y), \epsilon)$

    \tcc{Total loss and update}
    $\mathcal{L}_{\mathrm{total}} \gets \mathcal{L}_{\mathrm{data}} + \lambda_2 \mathcal{L}_{\mathrm{reg}} + \lambda_3 \mathcal{L}_{\mathrm{diff}} + \lambda_4 \mathcal{L}_{\mathrm{cls}}$ 
    Update $\theta$ using $\mathcal{L}_{\mathrm{total}}$; update $\phi'$ using $\mathcal{L}_{\mathrm{diff}}$
}

\KwOut{Trained generator $G_{\theta}$}
\end{algorithm*}

\subsection{Implement Details} 
\vspace{-0.1cm}
\noindent
\textbf{Training:}
The training takes over 30k iterations, with a batch size of 4 and a learning rate of $2\text{e}^{-5}$.
We adopted LoRA for fine-tuning. Follow the original setting~\cite{wu2024one,zhang2024degradation-s3diff}, for RCOD$_\text{S}$, the rank parameter in LoRA is set as $16$ for the VAE encoder and $32$ for the diffusion UNet, respectively. For RCOD$_\text{O}$, all rank is set to $4$.

\noindent
\textbf{Hardware and Software Environments:}
We train and test our model on NVIDIA A100 GPU. The amount of training memory is about 28 GB with batch size 1 using  RCOD$_\text{O}$. We use a Linux system with pytorch version 2.4.

\subsection{Experiments}
\subsubsection{Quantitative Comparisons:}
We quantitatively compare recent SOTA SR methods in Table \ref{tab:main_3data_SM} including synthetic and real-world data, multi-step diffusion, and GAN-based methods.
Compared with multi-step diffusion methods, our methods demonstrate better NR metrics and perceptual FR metrics across three datasets. Furthermore, their FR fidelity metrics are more competitive with those of multi-step diffusion methods when compared to other OSD methods.
On the DrealSR dataset, RCOD$_\text{O}$-Fid. can even surpass some multi-step diffusion methods in many FR metrics (PSNR, SSIM, and LPIPS) and NR metrics (MUSIQ, MANIQA, and CLIPIQA).
 
\begin{table*}[htbp]
  \centering
  
  \resizebox{0.86\textwidth}{!}{
\setlength{\tabcolsep}{3pt}

\begin{tabular}{c|c|c|cccccccc}
\toprule
Datasets & Type  & Methods &  PSNR↑  &  SSIM↑  &  LPIPS↓  &  DISTS↓  &  NIQE↓  &  MUSIQ↑  &  MANIQA↑  &  CLIPIQA↑  \\
\midrule
\multirow{20}[6]{*}{DIV2K-Val} & \multirow{2}[2]{*}{Non-Diff.} & BSRGAN &{24.58} &{0.6269} &{0.3351} &{0.2275} &{4.752} &{61.20} &{0.5071} &{0.5247} \\
      &       & Real-ESRGAN &{24.29} &{0.6371} &{0.3112} &{0.2141} &{4.679} &{61.06} &{0.5501} &{0.5277} \\
\cmidrule{2-11}      & \multirow{5}[2]{*}{Multi-step Diff.} & StableSR-s200 &{23.26} &{0.5726} &{\textcolor[rgb]{ 1,  0,  0}{\textbf{0.3113}}} &{\textcolor[rgb]{ 0,  .439,  .753}{0.2048}} &{4.758} &{65.92} &{0.6192} &{0.6771} \\
      &       & DiffBIR-s50 &{23.64} &{0.5647} &{0.3524} &{0.2128} &{\textcolor[rgb]{ 0,  .439,  .753}{4.704}} &{65.81} &{0.6210} &{0.6704} \\
      &       & SeeSR-s50 &{\textcolor[rgb]{ 0,  .439,  .753}{23.68}} &{\textcolor[rgb]{ 0,  .439,  .753}{0.6043}} &{\textcolor[rgb]{ 0,  .439,  .753}{0.3194}} &{\textcolor[rgb]{ 1,  0,  0}{\textbf{0.1968}}} &{4.810} &{\textcolor[rgb]{ 0,  .439,  .753}{68.67}} &{\textcolor[rgb]{ 0,  .439,  .753}{0.6240}} &{\textcolor[rgb]{ 1,  0,  0}{\textbf{0.6936}}} \\
      &       & PASD-s20 &{23.14} &{0.5505} &{0.3571} &{0.2207} &{\textcolor[rgb]{ 1,  0,  0}{\textbf{4.362}}} &{\textcolor[rgb]{ 1,  0,  0}{\textbf{68.95}}} &{\textcolor[rgb]{ 1,  0,  0}{\textbf{0.6483}}} &{\textcolor[rgb]{ 0,  .439,  .753}{0.6788}} \\
      &       & ResShift-s15 &{\textcolor[rgb]{ 1,  0,  0}{\textbf{24.65}}} &{\textcolor[rgb]{ 1,  0,  0}{\textbf{0.6181}}} &{0.3349} &{0.2213} &{6.821} &{61.09} &{0.5454} &{0.6071} \\
\cmidrule{2-11}      & \multirow{13}[2]{*}{One-step Diff.} & SinSR &{\textcolor[rgb]{ 0,  .439,  .753}{24.41}} &{0.6018} &{0.3240} &{0.2066} &{6.016} &{62.82} &{0.5386} &{0.6471} \\
      &       & PiSA-SR &{23.87} &{0.6058} &{0.2823} &{0.1934} &{4.550} &{69.68} &{0.6400} &{0.6927} \\
      &       & TSD-SR &{22.17} &{0.5602} &{0.2736} &{\textcolor[rgb]{ 0,  .439,  .753}{0.1855}} &{4.340} &{\textcolor[rgb]{ 0,  .439,  .753}{70.65}} &{0.6077} &{0.7148} \\
      &       & InvSR &{22.90} &{0.5910} &{0.5910} &{0.2058} &{4.689} &{68.90} &{0.6402} &{0.7181} \\
      &       & OSEDiff &{23.72} &{\textcolor[rgb]{ 0,  .439,  .753}{0.6108}} &{0.2941} &{0.1976} &{4.710} &{67.97} &{0.6148} &{0.6683} \\
      &       & \cellcolor[rgb]{ .929,  .929,  .929}RCOD$_\text{O}$-Fid. &{\cellcolor[rgb]{ .929,  .929,  .929}\textcolor[rgb]{ 1,  0,  0}{\textbf{24.42}}} &{\cellcolor[rgb]{ .929,  .929,  .929}\textcolor[rgb]{ 1,  0,  0}{\textbf{0.6250}}} &{\cellcolor[rgb]{ .929,  .929,  .929}0.2858} &{\cellcolor[rgb]{ .929,  .929,  .929}0.1969} &{\cellcolor[rgb]{ .929,  .929,  .929}5.041} &{\cellcolor[rgb]{ .929,  .929,  .929}69.50} &{\cellcolor[rgb]{ .929,  .929,  .929}\textcolor[rgb]{ 0,  .439,  .753}{0.6433}} &{\cellcolor[rgb]{ .929,  .929,  .929}0.6881} \\
      &       & \cellcolor[rgb]{ .929,  .929,  .929}RCOD$_\text{O}$-Neu. &{\cellcolor[rgb]{ .929,  .929,  .929}24.02} &{\cellcolor[rgb]{ .929,  .929,  .929}0.6147} &{\cellcolor[rgb]{ .929,  .929,  .929}0.2986} &{\cellcolor[rgb]{ .929,  .929,  .929}0.2057} &{\cellcolor[rgb]{ .929,  .929,  .929}4.780} &{\cellcolor[rgb]{ .929,  .929,  .929}70.21} &{\cellcolor[rgb]{ .929,  .929,  .929}\textcolor[rgb]{ 1,  0,  0}{\textbf{0.6478}}} &{\cellcolor[rgb]{ .929,  .929,  .929}0.7124} \\
      &       & \cellcolor[rgb]{ .929,  .929,  .929}RCOD$_\text{O}$-Real. &{\cellcolor[rgb]{ .929,  .929,  .929}23.52} &{\cellcolor[rgb]{ .929,  .929,  .929}0.5979} &{\cellcolor[rgb]{ .929,  .929,  .929}0.3285} &{\cellcolor[rgb]{ .929,  .929,  .929}0.2249} &{\cellcolor[rgb]{ .929,  .929,  .929}4.655} &{\cellcolor[rgb]{ .929,  .929,  .929}70.39} &{\cellcolor[rgb]{ .929,  .929,  .929}0.6332} &{\cellcolor[rgb]{ .929,  .929,  .929}\textcolor[rgb]{ 0,  .439,  .753}{0.7357}} \\
      &       & S3Diff &{23.53} &{0.5933} &{\textcolor[rgb]{ 0,  .439,  .753}{0.2581}} &{\textcolor[rgb]{ 1,  0,  0}{\textbf{0.1730}}} &{4.741} &{67.92} &{0.6328} &{0.7000} \\
      &       & \cellcolor[rgb]{ .929,  .929,  .929}RCOD$_\text{S}$-Fid. &{\cellcolor[rgb]{ .929,  .929,  .929}24.06} &{\cellcolor[rgb]{ .929,  .929,  .929}0.6103} &{\cellcolor[rgb]{ .929,  .929,  .929}\textcolor[rgb]{ 1,  0,  0}{\textbf{0.2543}}} &{\cellcolor[rgb]{ .929,  .929,  .929}0.1869} &{\cellcolor[rgb]{ .929,  .929,  .929}4.334} &{\cellcolor[rgb]{ .929,  .929,  .929}68.04} &{\cellcolor[rgb]{ .929,  .929,  .929}0.6188} &{\cellcolor[rgb]{ .929,  .929,  .929}0.6515} \\
      &       & \cellcolor[rgb]{ .929,  .929,  .929}RCOD$_\text{S}$-Neu. &{\cellcolor[rgb]{ .929,  .929,  .929}23.52} &{\cellcolor[rgb]{ .929,  .929,  .929}0.5858} &{\cellcolor[rgb]{ .929,  .929,  .929}0.2894} &{\cellcolor[rgb]{ .929,  .929,  .929}0.1895} &{\cellcolor[rgb]{ .929,  .929,  .929}\textcolor[rgb]{ 1,  0,  0}{\textbf{4.243}}} &{\cellcolor[rgb]{ .929,  .929,  .929}70.15} &{\cellcolor[rgb]{ .929,  .929,  .929}0.6358} &{\cellcolor[rgb]{ .929,  .929,  .929}0.7090} \\
      &       & \cellcolor[rgb]{ .929,  .929,  .929}RCOD$_\text{S}$-Real. &{\cellcolor[rgb]{ .929,  .929,  .929}23.01} &{\cellcolor[rgb]{ .929,  .929,  .929}0.5574} &{\cellcolor[rgb]{ .929,  .929,  .929}0.3078} &{\cellcolor[rgb]{ .929,  .929,  .929}0.2013} &{\cellcolor[rgb]{ .929,  .929,  .929}4.353} &{\cellcolor[rgb]{ .929,  .929,  .929}\textcolor[rgb]{ 1,  0,  0}{\textbf{70.75}}} &{\cellcolor[rgb]{ .929,  .929,  .929}0.6376} &{\cellcolor[rgb]{ .929,  .929,  .929}\textcolor[rgb]{ 1,  0,  0}{\textbf{0.7467}}} \\
      &       & \cellcolor[rgb]{ .929,  .929,  .929}RCOD$_\text{S}$-Adap. &{\cellcolor[rgb]{ .929,  .929,  .929}23.72} &{\cellcolor[rgb]{ .929,  .929,  .929}0.5932} &{\cellcolor[rgb]{ .929,  .929,  .929}0.2769} &{\cellcolor[rgb]{ .929,  .929,  .929}0.1880} &{\cellcolor[rgb]{ .929,  .929,  .929}\textcolor[rgb]{ 0,  .439,  .753}{4.260}} &{\cellcolor[rgb]{ .929,  .929,  .929}69.54} &{\cellcolor[rgb]{ .929,  .929,  .929}0.6308} &{\cellcolor[rgb]{ .929,  .929,  .929}0.6936} \\
\midrule
\midrule
      & \multirow{2}[2]{*}{Non-Diff.} & BSRGAN &{28.75} &{0.8031} &{0.2883} &{0.2142} &{6.519} &{57.14} &{0.4878} &{0.4915} \\
      &       & Real-ESRGAN &{28.64} &{0.8053} &{0.2847} &{0.2089} &{6.693} &{54.18} &{0.4907} &{0.4422} \\
\cmidrule{2-11}\multirow{18}[4]{*}{DrealSR} & \multirow{5}[2]{*}{Multi-step Diff.} & StableSR-s200 &{\textcolor[rgb]{ 0,  .439,  .753}{28.03}} &{0.7536} &{\textcolor[rgb]{ 0,  .439,  .753}{0.3284}} &{\textcolor[rgb]{ 1,  0,  0}{\textbf{0.2269}}} &{6.524} &{58.51} &{0.5601} &{0.6356} \\
      &       & DiffBIR-s50 &{26.71} &{0.6571} &{0.4557} &{0.2748} &{\textcolor[rgb]{ 0,  .439,  .753}{6.312}} &{61.07} &{0.5930} &{0.6395} \\
      &       & SeeSR-s50 &{28.17} &{\textcolor[rgb]{ 1,  0,  0}{\textbf{0.7691}}} &{\textcolor[rgb]{ 1,  0,  0}{\textbf{0.3189}}} &{\textcolor[rgb]{ 0,  .439,  .753}{0.2315}} &{6.397} &{\textcolor[rgb]{ 1,  0,  0}{\textbf{64.93}}} &{\textcolor[rgb]{ 0,  .439,  .753}{0.6042}} &{\textcolor[rgb]{ 0,  .439,  .753}{0.6804}} \\
      &       & PASD-s20 &{27.36} &{0.7073} &{0.3760} &{0.2531} &{\textcolor[rgb]{ 1,  0,  0}{\textbf{5.547}}} &{\textcolor[rgb]{ 0,  .439,  .753}{64.87}} &{\textcolor[rgb]{ 1,  0,  0}{\textbf{0.6169}}} &{\textcolor[rgb]{ 1,  0,  0}{\textbf{0.6808}}} \\
      &       & ResShift-s15 &{\textcolor[rgb]{ 1,  0,  0}{\textbf{28.46}}} &{\textcolor[rgb]{ 0,  .439,  .753}{0.7673}} &{0.4006} &{0.2656} &{8.125} &{50.60} &{0.4586} &{0.5342} \\
\cmidrule{2-11}      & \multirow{13}[2]{*}{One-step Diff.} & SinSR & \textcolor[rgb]{ 0,  .439,  .753}{28.36} & 0.7515  & 0.3665  & 0.2485  & 6.991  & 55.33  & 0.4884  & 0.6383  \\
      &       & PiSA-SR &{28.31} &{0.7804} &{0.2960} &{0.2169} &{6.200} &{66.11} &{0.6156} &{0.6970} \\
      &       & TSD-SR &{25.67} &{0.7132} &{0.3538} &{0.2459} &{5.991} &{65.99} &{0.6327} &{0.7137} \\
      &       & InvSR &{28.33} &{0.7502} &{0.3678} &{0.2481} &{6.941} &{55.27} &{0.4900} &{0.6385} \\
      &       & OSEDiff & 27.92  & \textcolor[rgb]{ 0,  .439,  .753}{0.7835} & \textcolor[rgb]{ 0,  .439,  .753}{0.2968} & 0.2165  & 6.490  & 64.65  & 0.5899  & 0.6963  \\
      &       & \cellcolor[rgb]{ .929,  .929,  .929}RCOD$_\text{O}$-Fid. & \cellcolor[rgb]{ .929,  .929,  .929}\textcolor[rgb]{ 1,  0,  0}{\textbf{28.90}} & \cellcolor[rgb]{ .929,  .929,  .929}\textcolor[rgb]{ 1,  0,  0}{\textbf{0.7906}} & \cellcolor[rgb]{ .929,  .929,  .929}\textcolor[rgb]{ 1,  0,  0}{\textbf{0.2919}} &{\cellcolor[rgb]{ .929,  .929,  .929}0.2186} &{\cellcolor[rgb]{ .929,  .929,  .929}6.817} &{\cellcolor[rgb]{ .929,  .929,  .929}66.72} &{\cellcolor[rgb]{ .929,  .929,  .929}0.6275} &{\cellcolor[rgb]{ .929,  .929,  .929}0.7023} \\
      &       & \cellcolor[rgb]{ .929,  .929,  .929}RCOD$_\text{O}$-Neu. &{\cellcolor[rgb]{ .929,  .929,  .929}28.30} &{\cellcolor[rgb]{ .929,  .929,  .929}0.7775} &{\cellcolor[rgb]{ .929,  .929,  .929}0.3080} &{\cellcolor[rgb]{ .929,  .929,  .929}0.2306} &{\cellcolor[rgb]{ .929,  .929,  .929}6.469} & \cellcolor[rgb]{ .929,  .929,  .929}\textcolor[rgb]{ 0,  .439,  .753}{68.03} & \cellcolor[rgb]{ .929,  .929,  .929}\textcolor[rgb]{ 1,  0,  0}{\textbf{0.6385}} &{\cellcolor[rgb]{ .929,  .929,  .929}0.7179} \\
      &       & \cellcolor[rgb]{ .929,  .929,  .929}RCOD$_\text{O}$-Real. &{\cellcolor[rgb]{ .929,  .929,  .929}27.59} &{\cellcolor[rgb]{ .929,  .929,  .929}0.7600} &{\cellcolor[rgb]{ .929,  .929,  .929}0.3389} &{\cellcolor[rgb]{ .929,  .929,  .929}0.2499} &{\cellcolor[rgb]{ .929,  .929,  .929}6.172} & \cellcolor[rgb]{ .929,  .929,  .929}\textcolor[rgb]{ 1,  0,  0}{\textbf{68.19}} &{\cellcolor[rgb]{ .929,  .929,  .929}0.6295} &{\cellcolor[rgb]{ .929,  .929,  .929}0.7325} \\
      &       & S3Diff & 27.54  & 0.7491  & 0.3109  & \textcolor[rgb]{ 1,  0,  0}{\textbf{0.2099}} & 6.212  & 63.94  & 0.6134  & 0.7130  \\
      &       & \cellcolor[rgb]{ .929,  .929,  .929}RCOD$_\text{S}$-Fid. & \cellcolor[rgb]{ .929,  .929,  .929}28.09  & \cellcolor[rgb]{ .929,  .929,  .929}0.7800  & \cellcolor[rgb]{ .929,  .929,  .929}0.3002  & \cellcolor[rgb]{ .929,  .929,  .929}\textcolor[rgb]{ 0,  .439,  .753}{0.2149} & \cellcolor[rgb]{ .929,  .929,  .929}5.871  & \cellcolor[rgb]{ .929,  .929,  .929}65.74  & \cellcolor[rgb]{ .929,  .929,  .929}0.6174  & \cellcolor[rgb]{ .929,  .929,  .929}0.6963  \\
      &       & \cellcolor[rgb]{ .929,  .929,  .929}RCOD$_\text{S}$-Neu. & \cellcolor[rgb]{ .929,  .929,  .929}27.48  & \cellcolor[rgb]{ .929,  .929,  .929}0.7526  & \cellcolor[rgb]{ .929,  .929,  .929}0.3256  & \cellcolor[rgb]{ .929,  .929,  .929}0.2251  & \cellcolor[rgb]{ .929,  .929,  .929}\textcolor[rgb]{ 0,  .439,  .753}{5.636} &{\cellcolor[rgb]{ .929,  .929,  .929}67.40} & \cellcolor[rgb]{ .929,  .929,  .929}\textcolor[rgb]{ 0,  .439,  .753}{0.6339} & \cellcolor[rgb]{ .929,  .929,  .929}\textcolor[rgb]{ 0,  .439,  .753}{0.7278} \\
      &       & \cellcolor[rgb]{ .929,  .929,  .929}RCOD$_\text{S}$-Real. & \cellcolor[rgb]{ .929,  .929,  .929}26.95  & \cellcolor[rgb]{ .929,  .929,  .929}0.7190  & \cellcolor[rgb]{ .929,  .929,  .929}0.3607  & \cellcolor[rgb]{ .929,  .929,  .929}0.2414  & \cellcolor[rgb]{ .929,  .929,  .929}\textcolor[rgb]{ 1,  0,  0}{\textbf{5.448}} &{\cellcolor[rgb]{ .929,  .929,  .929}67.58} & \cellcolor[rgb]{ .929,  .929,  .929}0.6317  & \cellcolor[rgb]{ .929,  .929,  .929}\textcolor[rgb]{ 1,  0,  0}{\textbf{0.7478}} \\
      &       & \cellcolor[rgb]{ .929,  .929,  .929}RCOD$_\text{S}$-Adap. & \cellcolor[rgb]{ .929,  .929,  .929}27.83  & \cellcolor[rgb]{ .929,  .929,  .929}0.7661  & \cellcolor[rgb]{ .929,  .929,  .929}0.3098  & \cellcolor[rgb]{ .929,  .929,  .929}0.2181  & \cellcolor[rgb]{ .929,  .929,  .929}5.768  & \cellcolor[rgb]{ .929,  .929,  .929}66.32  & \cellcolor[rgb]{ .929,  .929,  .929}0.6223  & \cellcolor[rgb]{ .929,  .929,  .929}0.7110  \\
\midrule
\midrule
      & \multirow{2}[2]{*}{Non-Diff.} & BSRGAN &{26.39} &{0.7654} &{0.2670} &{0.2121} &{5.657} &{63.21} &{0.5399} &{0.5001} \\
      &       & Real-ESRGAN &{25.69} &{0.7616} &{0.2727} &{0.2063} &{5.830} &{60.18} &{0.5487} &{0.4449} \\
\cmidrule{2-11}\multirow{18}[4]{*}{RealSR} & \multirow{5}[2]{*}{Multi-step Diff.} & StableSR-s200 &{24.70} &{0.7085} &{\textcolor[rgb]{ 0,  .439,  .753}{0.3018}} &{0.2288} &{5.912} &{65.78} &{0.6221} &{0.6178} \\
      &       & DiffBIR-s50 &{24.75} &{0.6567} &{0.3636} &{0.2312} &{5.535} &{64.98} &{0.6246} &{\textcolor[rgb]{ 0,  .439,  .753}{0.6463}} \\
      &       & SeeSR-s50 &{25.18} &{\textcolor[rgb]{ 0,  .439,  .753}{0.7216}} &{\textcolor[rgb]{ 1,  0,  0}{\textbf{0.3009}}} &{\textcolor[rgb]{ 1,  0,  0}{\textbf{0.2223}}} &{\textcolor[rgb]{ 1,  0,  0}{\textbf{5.408}}} &{\textcolor[rgb]{ 1,  0,  0}{\textbf{69.77}}} &{\textcolor[rgb]{ 0,  .439,  .753}{0.6442}} &{0.6612} \\
      &       & PASD-s20 &{\textcolor[rgb]{ 0,  .439,  .753}{25.21}} &{0.6798} &{0.3380} &{\textcolor[rgb]{ 0,  .439,  .753}{0.2260}} &{\textcolor[rgb]{ 0,  .439,  .753}{5.414}} &{\textcolor[rgb]{ 0,  .439,  .753}{68.75}} &{\textcolor[rgb]{ 1,  0,  0}{\textbf{0.6487}}} &{\textcolor[rgb]{ 1,  0,  0}{\textbf{0.6620}}} \\
      &       & ResShift-s15 &{\textcolor[rgb]{ 1,  0,  0}{\textbf{26.31}}} &{\textcolor[rgb]{ 1,  0,  0}{\textbf{0.7421}}} &{0.3460} &{0.2498} &{7.264} &{58.43} &{0.5285} &{0.5444} \\
\cmidrule{2-11}      & \multirow{13}[2]{*}{One-step Diff.} & SinSR & \textcolor[rgb]{ 1,  0,  0}{\textbf{26.28}} & 0.7347  & 0.3188  & 0.2353  & 6.287  & 60.80  & 0.5385  & 0.6122  \\
      &       & PiSA-SR &{25.50} &{0.7417} & \textcolor[rgb]{ 0,  .439,  .753}{0.2672} &{0.2044} &{5.500} &{70.15} &{0.6560} &{0.6702} \\
      &       & TSD-SR &{23.41} &{0.6886} &{0.2805} &{0.2183} &{5.093} & \textcolor[rgb]{ 1,  0,  0}{\textbf{70.77}} &{0.6311} & \textcolor[rgb]{ 1,  0,  0}{\textbf{0.7193}} \\
      &       & InvSR &{24.13} &{0.7125} &{0.2871} &{0.2123} &{5.626} &{68.54} &{0.6619} &{0.6790} \\
      &       & OSEDiff & 25.15  & 0.7341  & 0.2921  & 0.2128  & 5.648  & 69.09  & 0.6326  & 0.6693  \\
      &       & \cellcolor[rgb]{ .929,  .929,  .929}RCOD$_\text{O}$-Fid. & \cellcolor[rgb]{ .929,  .929,  .929}\textcolor[rgb]{ 0,  .439,  .753}{26.01} & \cellcolor[rgb]{ .929,  .929,  .929}\textcolor[rgb]{ 1,  0,  0}{\textbf{0.7427}} &{\cellcolor[rgb]{ .929,  .929,  .929}0.2796} &{\cellcolor[rgb]{ .929,  .929,  .929}0.2103} &{\cellcolor[rgb]{ .929,  .929,  .929}5.911} &{\cellcolor[rgb]{ .929,  .929,  .929}70.25} &{\cellcolor[rgb]{ .929,  .929,  .929}0.6647} &{\cellcolor[rgb]{ .929,  .929,  .929}0.6866} \\
      &       & \cellcolor[rgb]{ .929,  .929,  .929}RCOD$_\text{O}$-Neu. &{\cellcolor[rgb]{ .929,  .929,  .929}25.39} &{\cellcolor[rgb]{ .929,  .929,  .929}0.7264} &{\cellcolor[rgb]{ .929,  .929,  .929}0.2939} &{\cellcolor[rgb]{ .929,  .929,  .929}0.2190} &{\cellcolor[rgb]{ .929,  .929,  .929}5.497} &{\cellcolor[rgb]{ .929,  .929,  .929}70.34} & \cellcolor[rgb]{ .929,  .929,  .929}\textcolor[rgb]{ 0,  .439,  .753}{0.6750} &{\cellcolor[rgb]{ .929,  .929,  .929}0.7022} \\
      &       & \cellcolor[rgb]{ .929,  .929,  .929}RCOD$_\text{O}$-Real. &{\cellcolor[rgb]{ .929,  .929,  .929}24.62} &{\cellcolor[rgb]{ .929,  .929,  .929}0.7011} &{\cellcolor[rgb]{ .929,  .929,  .929}0.3296} &{\cellcolor[rgb]{ .929,  .929,  .929}0.2375} &{\cellcolor[rgb]{ .929,  .929,  .929}5.341} & \cellcolor[rgb]{ .929,  .929,  .929}\textcolor[rgb]{ 0,  .439,  .753}{70.76} &{\cellcolor[rgb]{ .929,  .929,  .929}0.6650} &{\cellcolor[rgb]{ .929,  .929,  .929}0.7084} \\
      &       & S3Diff & 25.18  & 0.7269  & 0.2721  & \textcolor[rgb]{ 0,  .439,  .753}{0.2005} & 5.269  & 67.82  & 0.6437  & 0.6727  \\
      &       & \cellcolor[rgb]{ .929,  .929,  .929}RCOD$_\text{S}$-Fid. & \cellcolor[rgb]{ .929,  .929,  .929}25.42  & \cellcolor[rgb]{ .929,  .929,  .929}\textcolor[rgb]{ 0,  .439,  .753}{0.7392} & \cellcolor[rgb]{ .929,  .929,  .929}\textcolor[rgb]{ 1,  0,  0}{\textbf{0.2647}} & \cellcolor[rgb]{ .929,  .929,  .929}\textcolor[rgb]{ 1,  0,  0}{\textbf{0.1976}} & \cellcolor[rgb]{ .929,  .929,  .929}5.095  & \cellcolor[rgb]{ .929,  .929,  .929}69.46  & \cellcolor[rgb]{ .929,  .929,  .929}0.6605  & \cellcolor[rgb]{ .929,  .929,  .929}0.6509  \\
      &       & \cellcolor[rgb]{ .929,  .929,  .929}RCOD$_\text{S}$-Neu. & \cellcolor[rgb]{ .929,  .929,  .929}24.78  & \cellcolor[rgb]{ .929,  .929,  .929}0.7130  & \cellcolor[rgb]{ .929,  .929,  .929}0.2855  & \cellcolor[rgb]{ .929,  .929,  .929}0.2073  & \cellcolor[rgb]{ .929,  .929,  .929}\textcolor[rgb]{ 0,  .439,  .753}{5.024} &{\cellcolor[rgb]{ .929,  .929,  .929}70.55} & \cellcolor[rgb]{ .929,  .929,  .929}\textcolor[rgb]{ 1,  0,  0}{\textbf{0.6757}} & \cellcolor[rgb]{ .929,  .929,  .929}0.6886  \\
      &       & \cellcolor[rgb]{ .929,  .929,  .929}RCOD$_\text{S}$-Real. & \cellcolor[rgb]{ .929,  .929,  .929}24.08  & \cellcolor[rgb]{ .929,  .929,  .929}0.6759  & \cellcolor[rgb]{ .929,  .929,  .929}0.3228  & \cellcolor[rgb]{ .929,  .929,  .929}0.2236  & \cellcolor[rgb]{ .929,  .929,  .929}\textcolor[rgb]{ 1,  0,  0}{\textbf{4.900}} &{\cellcolor[rgb]{ .929,  .929,  .929}70.65} & \cellcolor[rgb]{ .929,  .929,  .929}0.6719  & \cellcolor[rgb]{ .929,  .929,  .929}\textcolor[rgb]{ 0,  .439,  .753}{0.7086} \\
      &       & \cellcolor[rgb]{ .929,  .929,  .929}RCOD$_\text{S}$-Adap. & \cellcolor[rgb]{ .929,  .929,  .929}25.23  & \cellcolor[rgb]{ .929,  .929,  .929}0.7313  & \cellcolor[rgb]{ .929,  .929,  .929}0.2714  & \cellcolor[rgb]{ .929,  .929,  .929}0.2010  & \cellcolor[rgb]{ .929,  .929,  .929}5.033  & \cellcolor[rgb]{ .929,  .929,  .929}69.72  & \cellcolor[rgb]{ .929,  .929,  .929}0.6646  & \cellcolor[rgb]{ .929,  .929,  .929}0.6622  \\
\bottomrule
\bottomrule
\end{tabular}%
 }
 
   \caption{Quantitative comparison with state-of-the-art methods on both synthetic and real-world benchmarks. The best and second best results within both multi-step and one-step diffusion-based methods are highlighted in \textcolor{red}{\textbf{red}} and \textcolor{blue}{{blue}}, respectively. }
    \label{tab:main_3data_SM}%
\end{table*}%

\subsubsection{Qualitative Comparisons:}
\begin{figure*} 
  \centering
    \begin{tabular}{cc}
\hspace{-0.4cm}
\begin{adjustbox}{valign=t}
\begin{tabular}{c}
\includegraphics[width=0.382\textwidth]{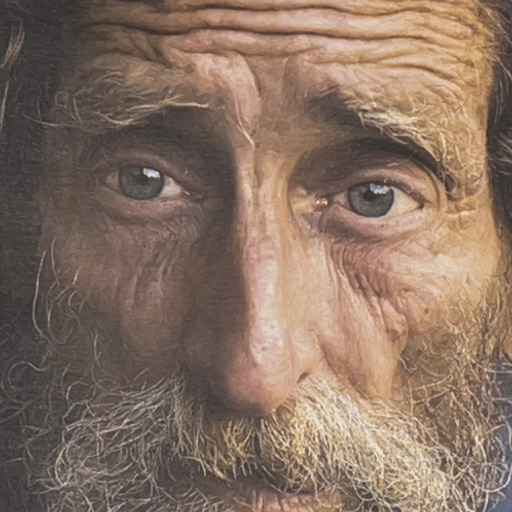} \\
Nikon\_047 HR
\end{tabular}
\end{adjustbox}
\hspace{-0.46cm}

\begin{adjustbox}{valign=t}
\begin{tabular}{cccc}
\includegraphics[width=0.118\textwidth]{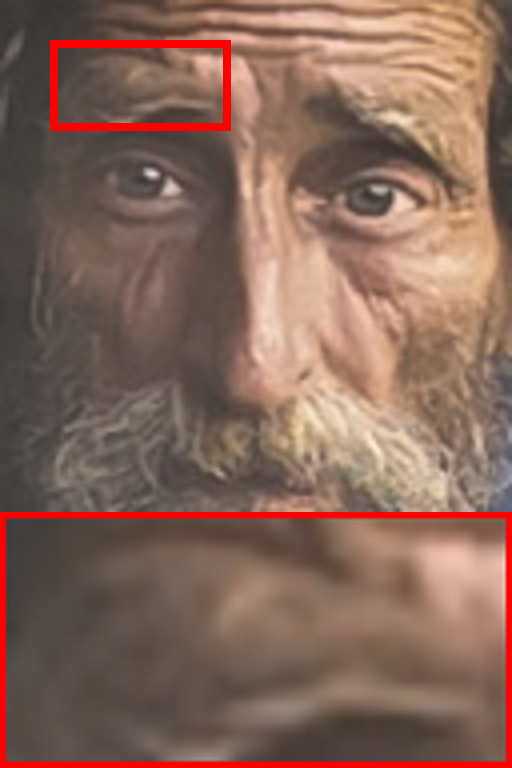} \hspace{-4mm} &
\includegraphics[width=0.118\textwidth]{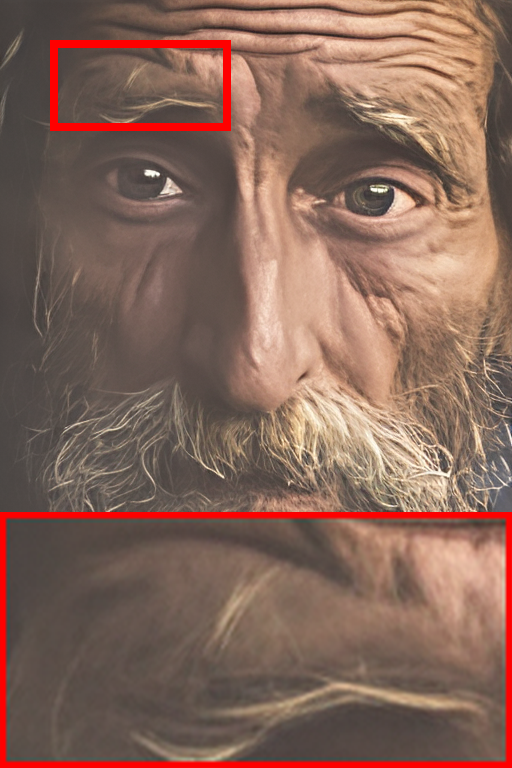} \hspace{-4mm} &
\includegraphics[width=0.118\textwidth]{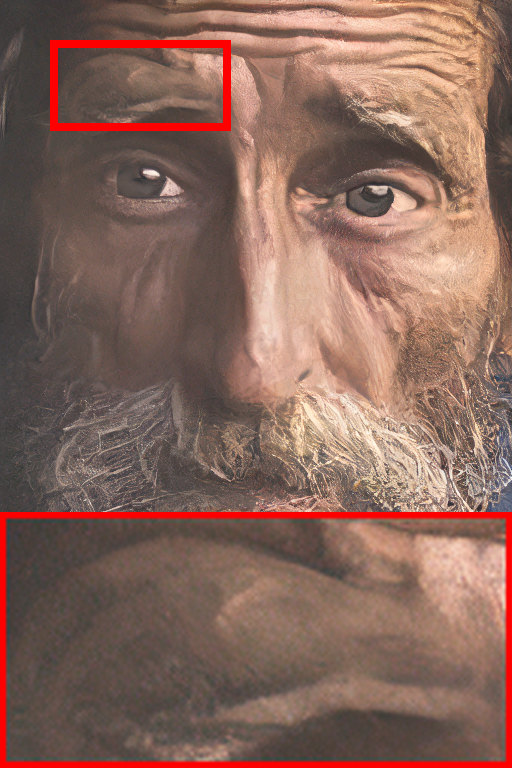} \hspace{-4mm} &
\includegraphics[width=0.118\textwidth]{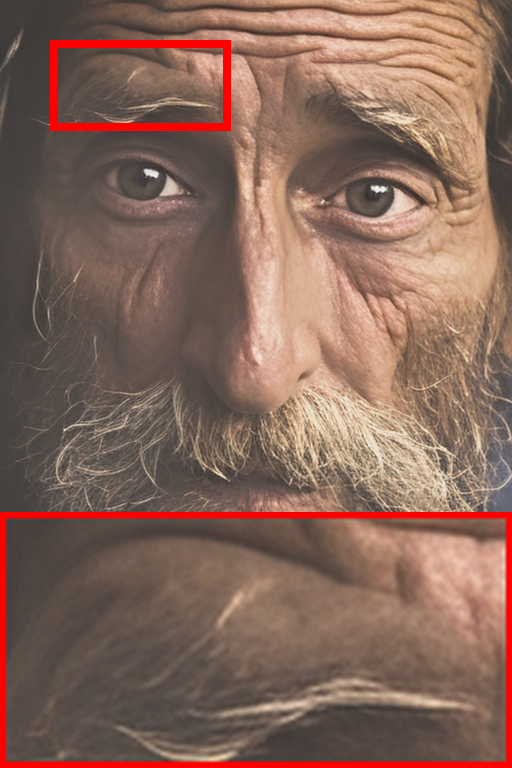} \hspace{-4mm} \\
LR \hspace{-4mm} &
StableSR \hspace{-4mm} &
SinSR \hspace{-4mm} &
PiSA-SR\hspace{-4mm} \\
\includegraphics[width=0.118\textwidth]{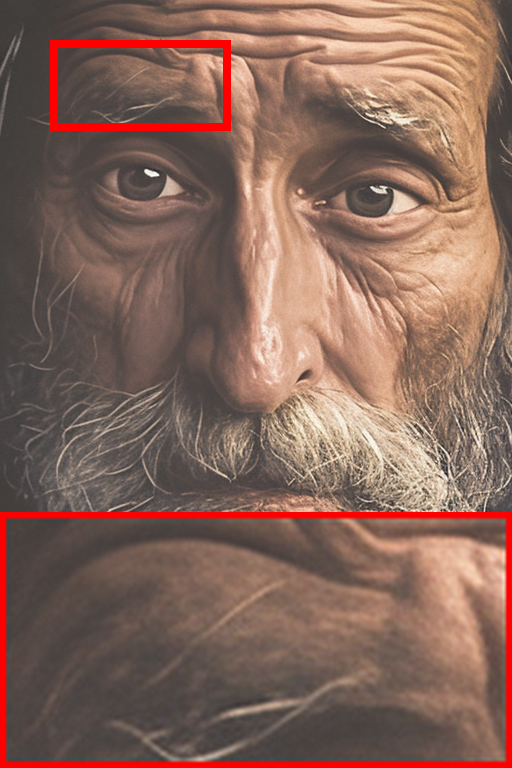} \hspace{-4mm} &
\includegraphics[width=0.118\textwidth]{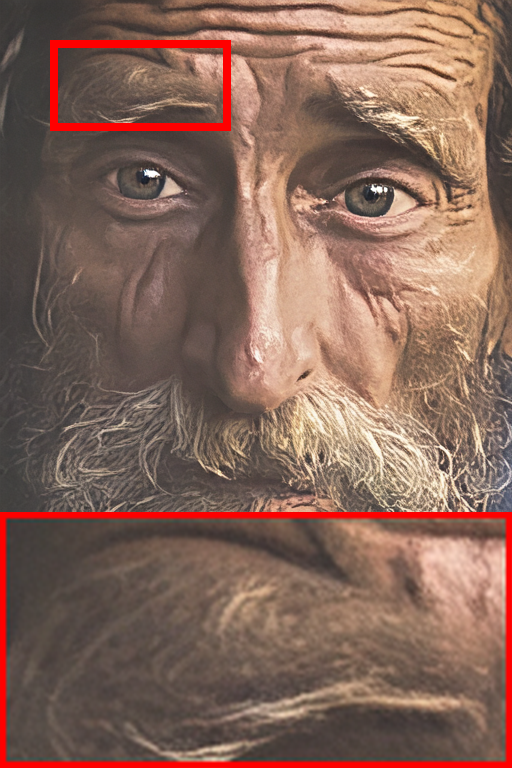} \hspace{-4mm} &
\includegraphics[width=0.118\textwidth]{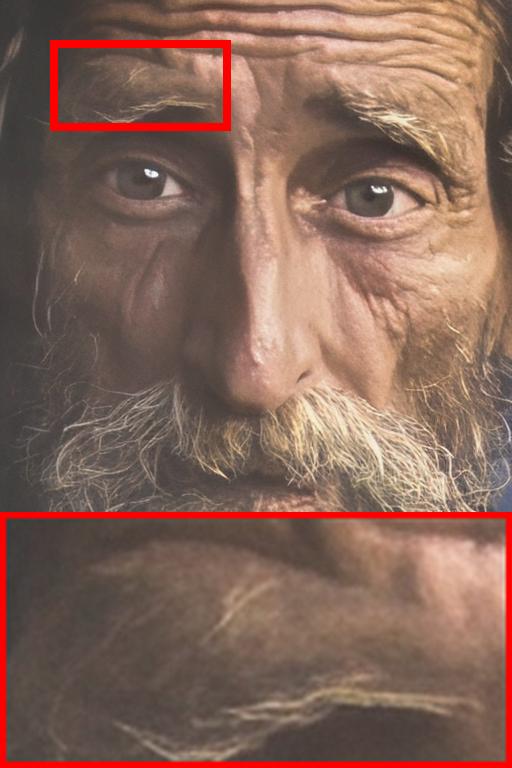} \hspace{-4mm} &
\includegraphics[width=0.118\textwidth]{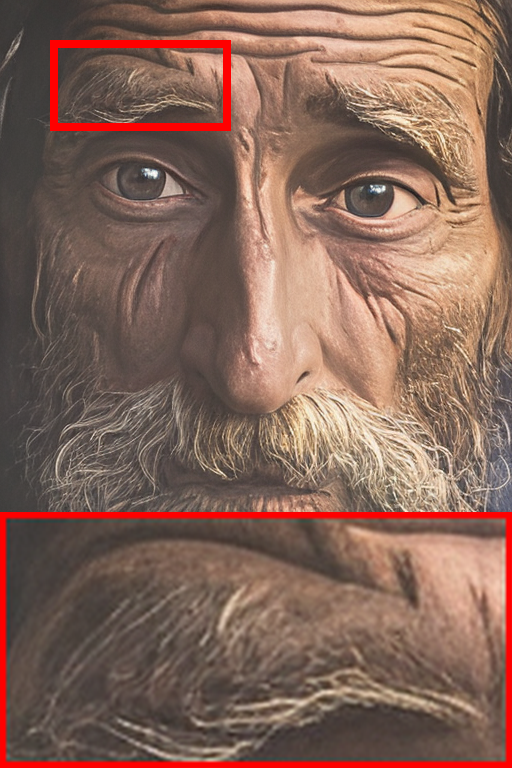} 
\hspace{-4mm} \\
InvSR \hspace{-4mm} &
TSD-SR\hspace{-4mm} &
OSEDiff \hspace{-4mm} &
RCOD-O.-Real. \hspace{-4mm} \\ 
\end{tabular}
\end{adjustbox}
\end{tabular}

    \begin{tabular}{cc}
\hspace{-0.4cm}
\begin{adjustbox}{valign=t}
\begin{tabular}{c}
\includegraphics[width=0.382\textwidth]{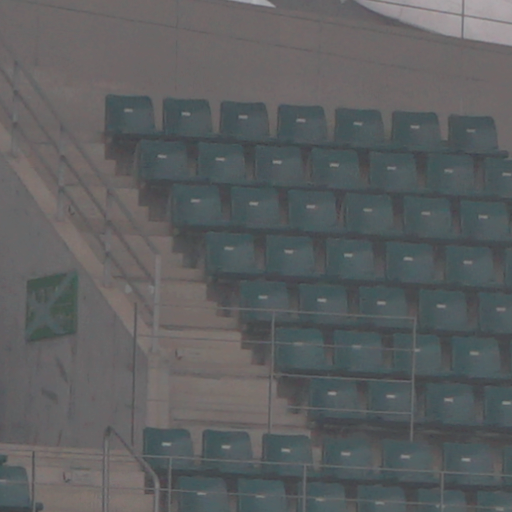} \\
Sony\_49 HR
\end{tabular}
\end{adjustbox}
\hspace{-0.46cm}

\begin{adjustbox}{valign=t}
\begin{tabular}{cccc}
\includegraphics[width=0.118\textwidth]{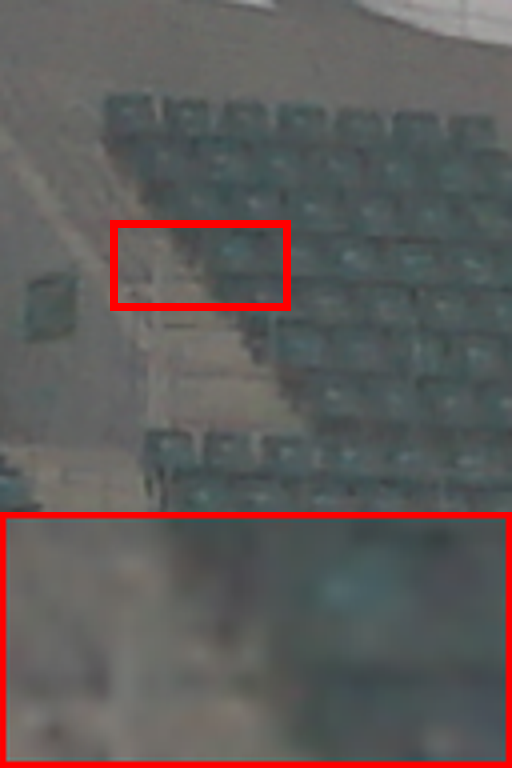} \hspace{-4mm} &
\includegraphics[width=0.118\textwidth]{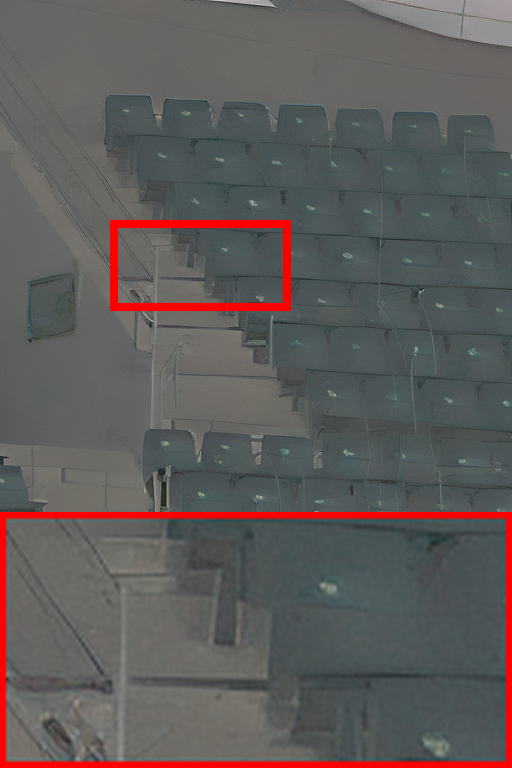} \hspace{-4mm} &
\includegraphics[width=0.118\textwidth]{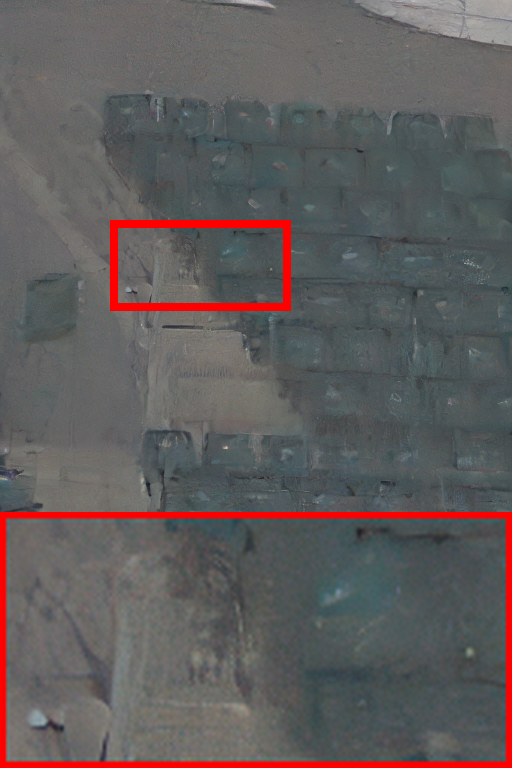} \hspace{-4mm} &
\includegraphics[width=0.118\textwidth]{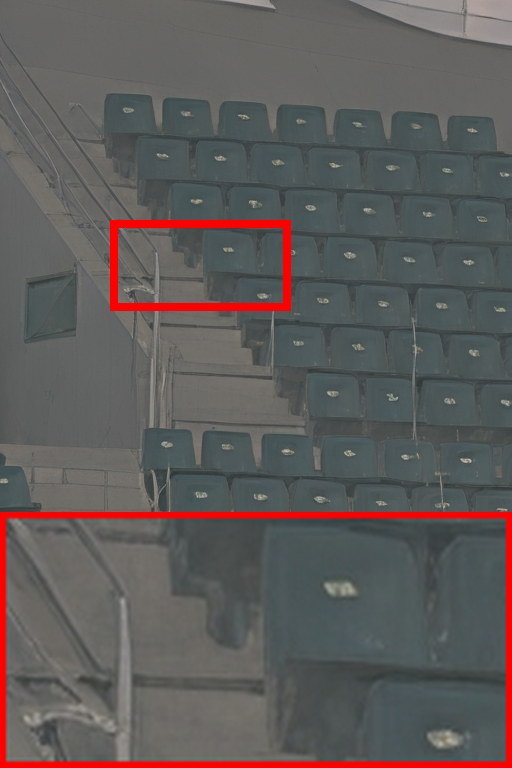} \hspace{-4mm} \\
LR \hspace{-4mm} &
StableSR \hspace{-4mm} &
SinSR \hspace{-4mm} &
PiSA-SR\hspace{-4mm} \\
\includegraphics[width=0.118\textwidth]{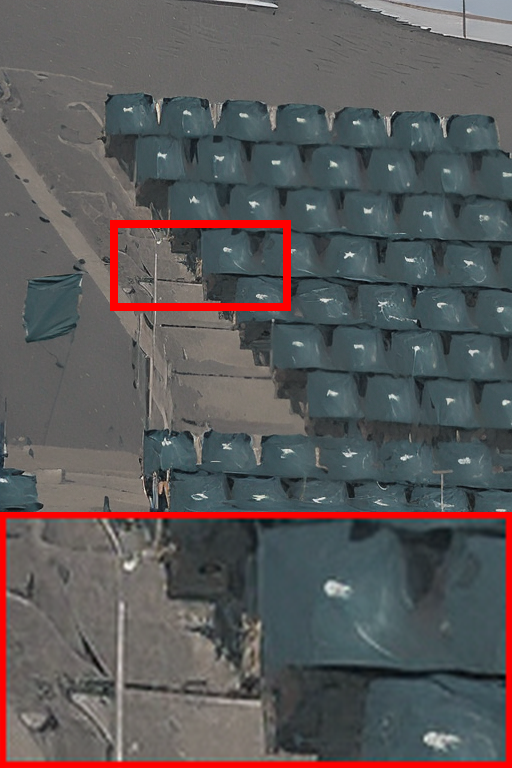} \hspace{-4mm} &
\includegraphics[width=0.118\textwidth]{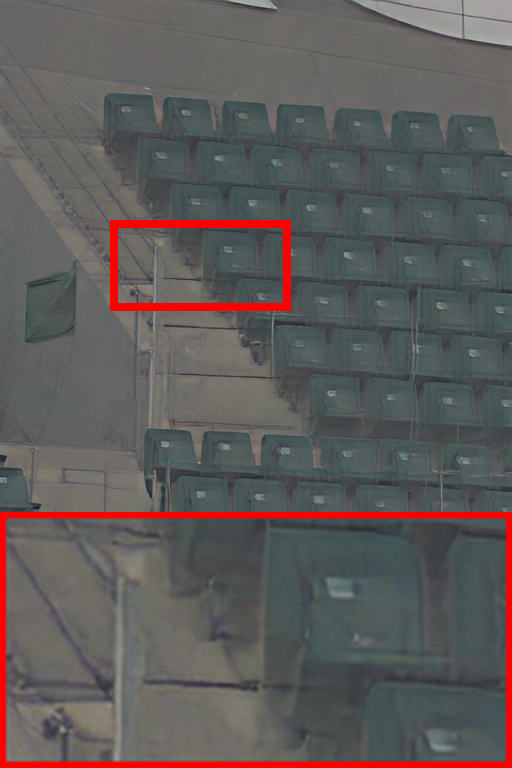} \hspace{-4mm} &
\includegraphics[width=0.118\textwidth]{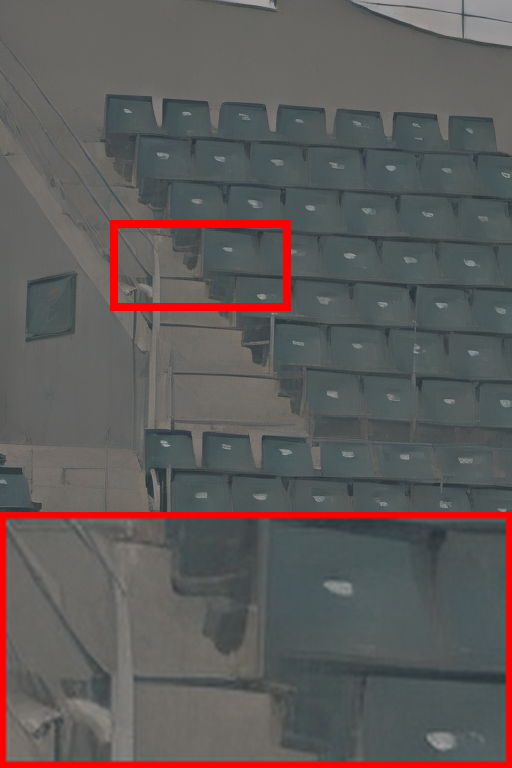} \hspace{-4mm} &
\includegraphics[width=0.118\textwidth]{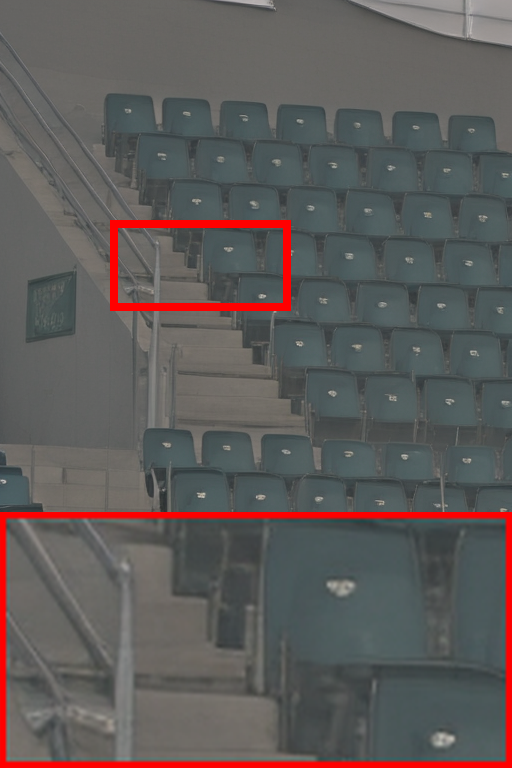} 
\hspace{-4mm} \\
InvSR \hspace{-4mm} &
TSD-SR\hspace{-4mm} &
OSEDiff \hspace{-4mm} &
RCOD-O.-Fid. \hspace{-4mm} \\ 
\end{tabular}
\end{adjustbox}
\end{tabular}

    \begin{tabular}{cc}
\hspace{-0.4cm}
\begin{adjustbox}{valign=t}
\begin{tabular}{c}
\includegraphics[width=0.382\textwidth]{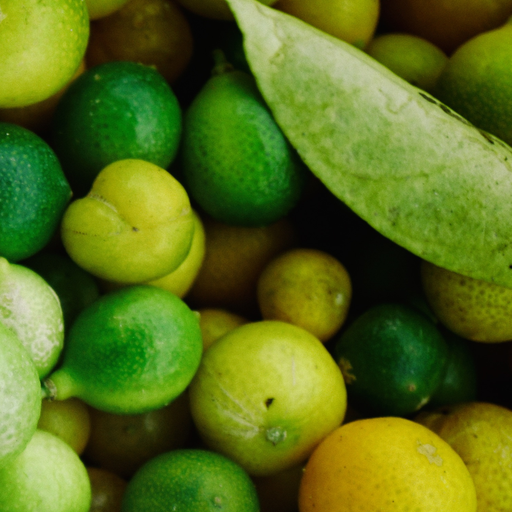} \\
0802\_pch\_00007 HR
\end{tabular}
\end{adjustbox}
\hspace{-0.46cm}

\begin{adjustbox}{valign=t}
\begin{tabular}{cccc}
\includegraphics[width=0.118\textwidth]{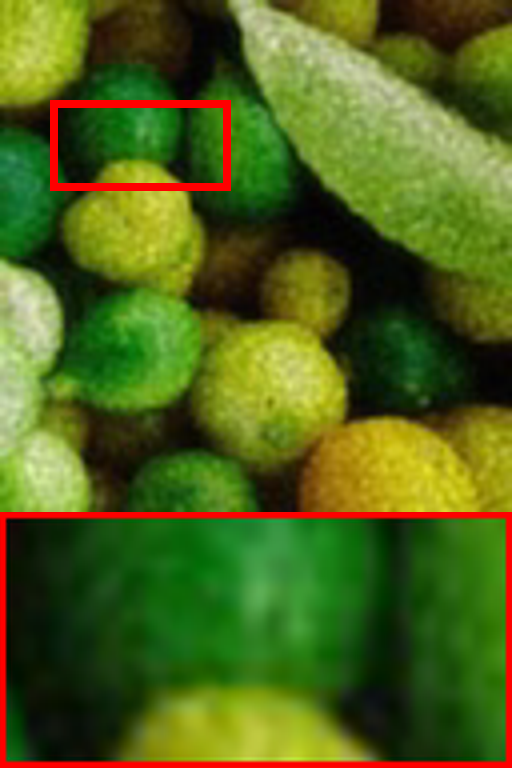} \hspace{-4mm} &
\includegraphics[width=0.118\textwidth]{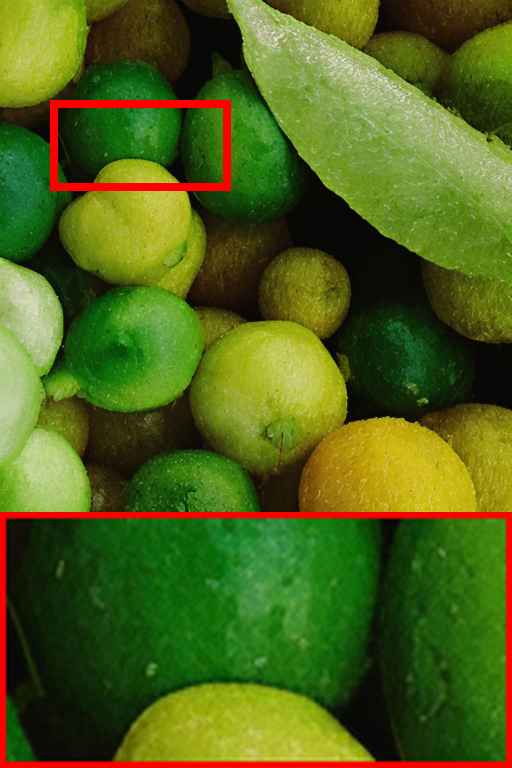} \hspace{-4mm} &
\includegraphics[width=0.118\textwidth]{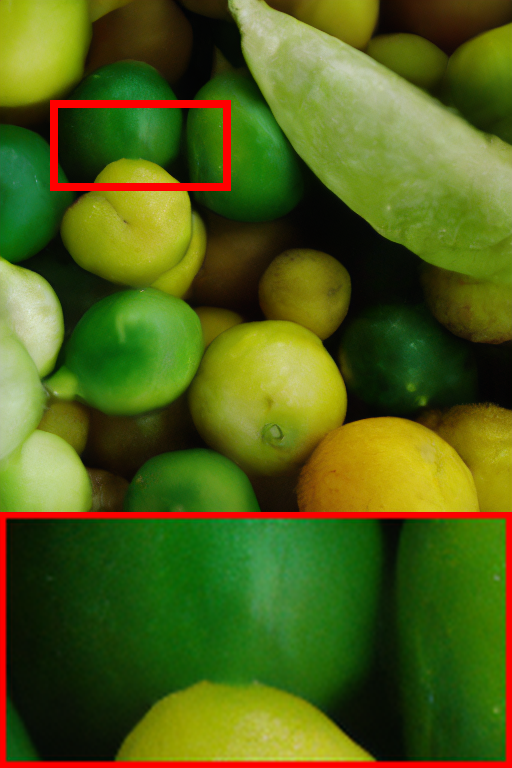} \hspace{-4mm} &
\includegraphics[width=0.118\textwidth]{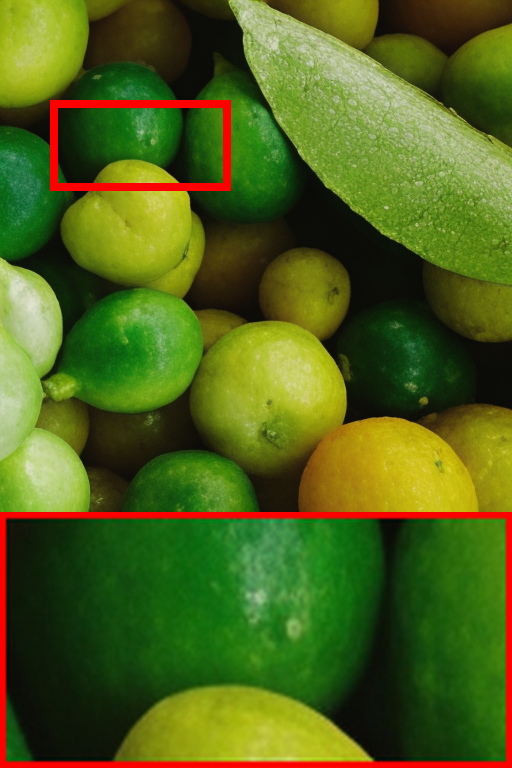} \hspace{-4mm} \\
LR \hspace{-4mm} &
StableSR \hspace{-4mm} &
SinSR \hspace{-4mm} &
PiSA-SR\hspace{-4mm} \\
\includegraphics[width=0.118\textwidth]{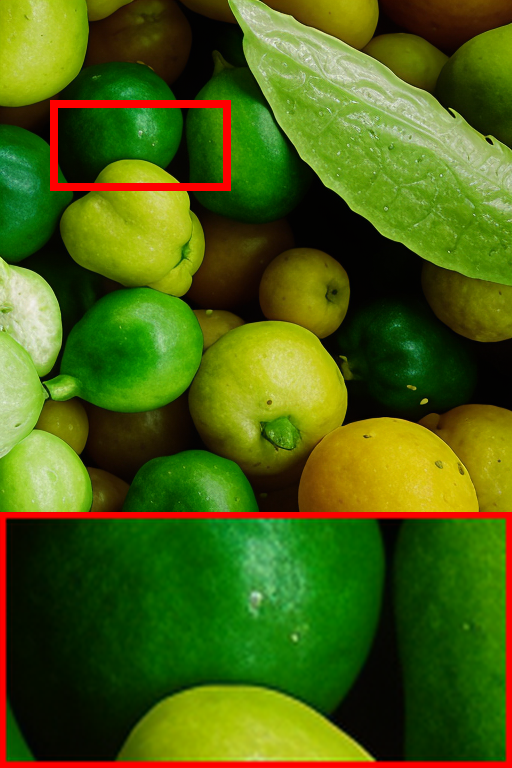} \hspace{-4mm} &
\includegraphics[width=0.118\textwidth]{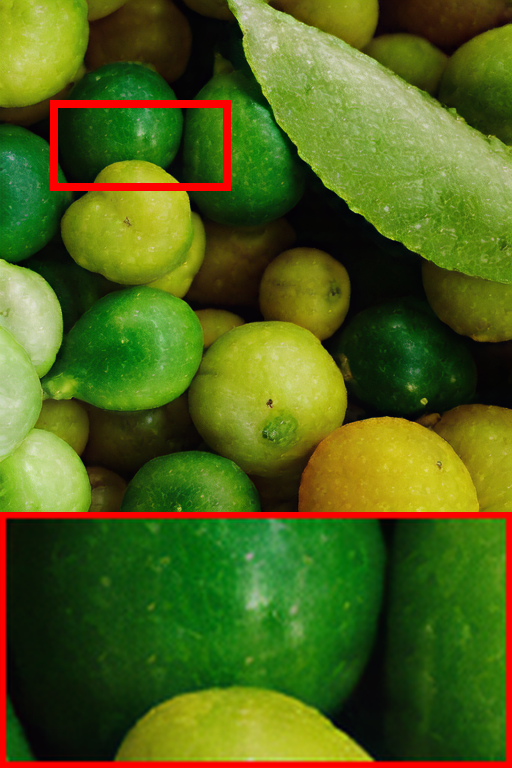} \hspace{-4mm} &
\includegraphics[width=0.118\textwidth]{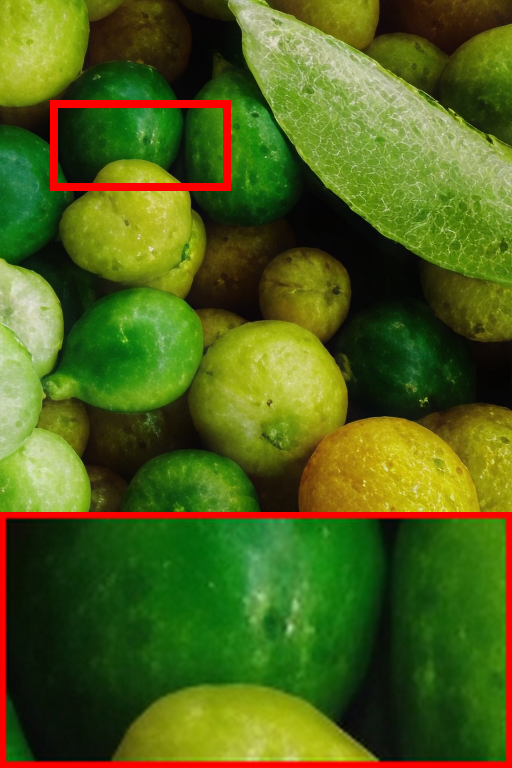} \hspace{-4mm} &
\includegraphics[width=0.118\textwidth]{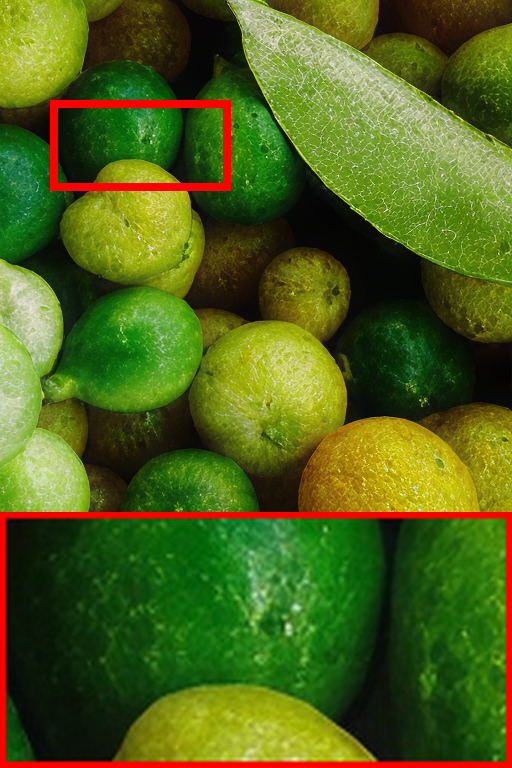} 
\hspace{-4mm} \\
InvSR \hspace{-4mm} &
TSD-SR\hspace{-4mm} &
S3Diff \hspace{-4mm} &
RCOD-S.-Real. \hspace{-4mm} \\ 
\end{tabular}
\end{adjustbox}
\end{tabular}
\vspace{-3mm}

\caption{Visual comparison ($\times$4) on RealSR (Nikon\_047), DRealSR (Sony\_49), and DIV2K (0802\_pch\_00007) data.}
\vspace{-4mm}
\label{fig:main_vis_compare_SM}
\end{figure*}

We conduct more qualitative comparisons in Fig.~\ref{fig:main_vis_compare_SM}. On Sony\_0049 (DRealSR), RCOD$_\text{O}$-Fid. shows fewer artefacts of handrail than other OSD methods. On Nikon\_047 (RealSR) and 0802\_pch\_00007 (DIV2K), RCOD$_\text{O}$-Real. and RCOD$_\text{S}$-Real. both generate more detailed textures than other OSD methods. 

\subsection{Ablation Study}
\textbf{Effectiveness of VPIM and DAS:} To validate the effectiveness of VPIM, we perform ablation studies by replacing the VLM and text encoder (``Text Model'') with VPIM in Table~\ref{tab:abla_module} for the ``w/o VPIM-Fid.'' and ``w/o VPIM-Real.'' configurations. We observe that both configurations achieve better PSNR, LPIPS, and MANIQA scores than the baseline (the original OSEDiff). This indicates that VPIM provides not only sufficient semantic information but also fidelity information from LR images. {
By removing DAS during distillation, as implemented in the w/o DAS-Fid.'' and w/o DAS-Real.'' configurations, we observe that DAS amplifies the performance gap between the fidelity-oriented and realism-oriented configurations. The fidelity-oriented variant achieves higher PSNR and lower LPIPS, while the realism-oriented variant yields higher MANIQA scores—indicating improved realism. This divergence in evaluation metrics demonstrates that DAS effectively enhances LDG's ability to decouple fidelity preservation from realism generation, enabling finer control over the fidelity–realism trade-off through adaptive timestep regularization.
}

\begin{figure}
\vspace{-1mm}
    \centering
    \includegraphics[width=1\linewidth, trim=0 0 0  0, clip]{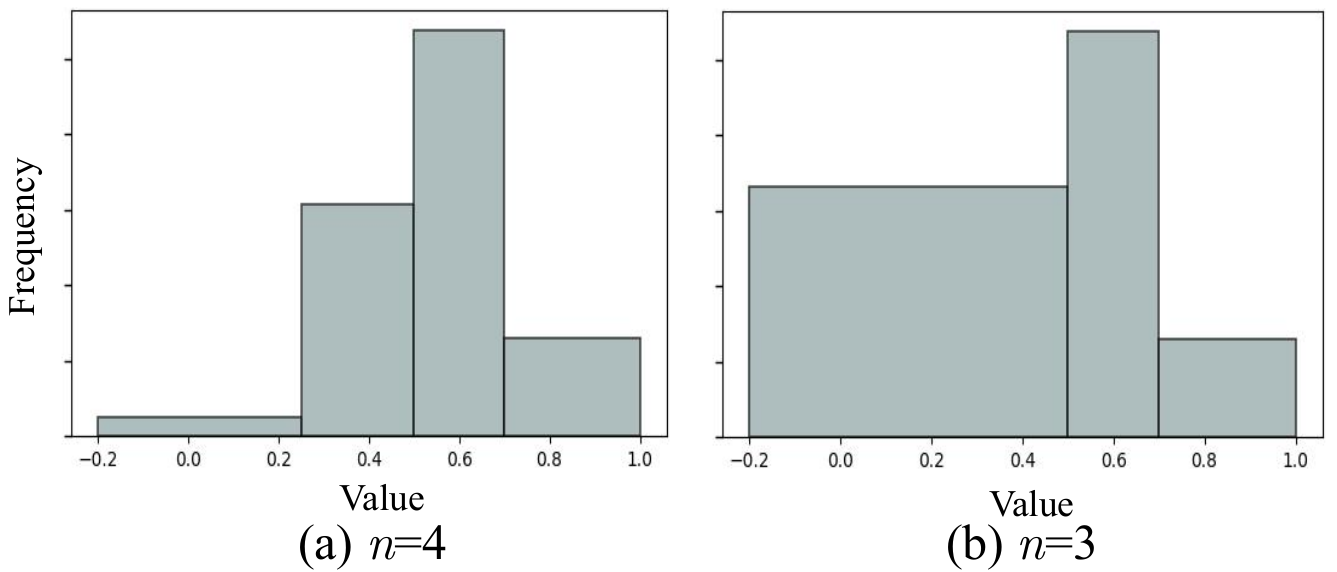}
    \vspace{-6mm}
    \caption{Distribution of (a) mean value of LR and HR training images in the latent domain, (b) the $M_L$ metric in the latent domain of VAE before and after training.}
        \vspace{-1mm}
    \label{fig:three_hist_SM}
\end{figure}

\noindent\textbf{Choice of $n$:} As mentioned in section ``Latent Domain Grouping'', $n$ can only be $\leq 4$. For sufficient flexibility, we consider $n \geq 3$. Therefore, we compare the training results for $n=3$ and $n=4$ in Table~\ref{tab:abla_step}. We find that RCOD$_\text{O}$-Real. ($n=4$) performs much worse than RCOD$_\text{O}$-Real. ($n=3$). To explain this phenomenon, we visualize the training data distribution for the two choices in Fig.~\ref{fig:three_hist_SM}. We observe that there is too little data for training at large $t$ (when $M_L$ is small). Thus, we finally choose $n=3$ for our experiment.

\noindent \textbf{Robustness of $M_L$:} Due to the encoder of the VAE being trainable in most SD-based method settings, we consider a domain shift phenomenon during training: the distribution of $M_L$ may change after training. Therefore, we compare the $M_L$ values of the entire training dataset before and after training in main manuscript Fig.~4 (b). The histogram shows a slight shift in the metric, with values becoming closer to larger similarities, but the overall distribution remains similar. This indicates that even though the model aligns the feature domains of LR and HR, $M_L$ retains its robustness in measuring the divergence in the latent domain.

\begin{figure*}[t]
    \centering
    \includegraphics[width=1\linewidth, trim=0 0 0  0, clip]{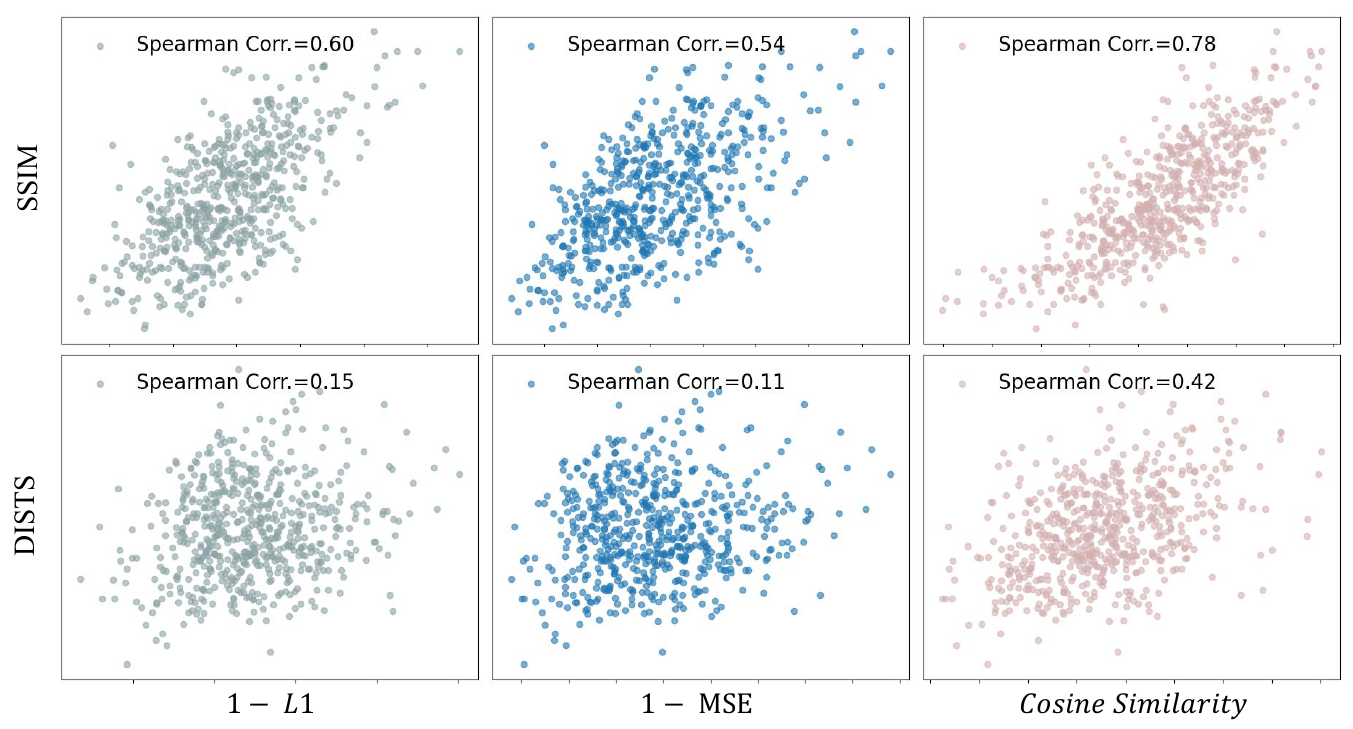}
    \vspace{-6mm}
    \caption{Visualization of $|\text{Spearman coefficient}|\uparrow$ of different $M_L$ distances and image quality metrics.}
        \vspace{-1mm}
    \label{fig:IQA_corr_SM}
\end{figure*}

\noindent\textbf{Correlation of $M_L$ and image quality:} In Table~1(b) in main manuscript, we calculate $|\text{Spearman coefficient}|\uparrow$ of different $M_L$ and image quality metrics (which are calculated by LR and HR images). Fig.~\ref{fig:IQA_corr_SM} visualize the correlation of randomly selected 600 points. We can find that cosine similarity exhibits a higher correlation coefficient with image quality metrics compared to other distances.

\begin{table*}
  \centering
      \resizebox{\linewidth}{!}{
 \begin{tabular}{c|ccccccc}
    \toprule
          & LDG & VPIM & Text Model & DAS & PSNR↑ & LPIPS↓ & MANIQA↑ \\
    \midrule
    Baseline & ×     & ×     & \checkmark & × & 25.15 & 0.2921 & 0.6326 \\
    w/o VPIM-Fid. & \checkmark & ×     & \checkmark & × & 25.99 & 0.2824 & 0.6541 \\
    w/o VPIM-Real. & \checkmark & ×     & \checkmark & × & 24.23 & 0.3242 & {0.6642} \\
    w/o LDG & ×     & \checkmark & ×     & × & 25.34 & 0.2877 & 0.6320 \\
    w/o DAS-Fid. & \checkmark & \checkmark & ×     & × & 25.90 & 0.2806 & 0.6553 \\
    w/o DAS-Real. & \checkmark & \checkmark & × & × & 24.89 & 0.3275 & 0.6602 \\
    \textbf{RCOD$_\text{O}$-Fid.} & \checkmark & \checkmark & ×     & \checkmark & \textbf{26.01} & \textbf{0.2796} & 0.6647 \\
    \textbf{RCOD$_\text{O}$-Real.} & \checkmark & \checkmark & ×     & \checkmark & 24.62 &0.3296 & \textbf{0.6650} \\
    \bottomrule
    \end{tabular}%
    }
    \caption{Ablation study of modules. The best results are highlighted in bold.}
      \label{tab:abla_module}%
\end{table*}%

\begin{table*}
  \centering
      \resizebox{1\linewidth}{!}{
    \begin{tabular}{l|cccc}
    \toprule
          & PSNR↑ & DISTS↓ & LPIPS↓ & MANIQA↑ \\
    \midrule
    RCOD$_\text{O}$-Fid. ($n=4$)& 26.00 & 0.2111 & 0.2805 & 0.6490 \\
    RCOD$_\text{O}$-Fid. ($n=3$) & \textbf{26.01} & \textbf{0.2103} & \textbf{0.2796} & 0.6647 \\
    RCOD$_\text{O}$-Real. ($n=4$) & 24.02 & 0.2590 & 0.3588 & 0.6400 \\
    RCOD$_\text{O}$-Real. ($n=3$) & 24.62 & 0.2375 & 0.3296 &  \textbf{0.6650} \\
    \bottomrule
    \end{tabular}%
    }
 \caption{Ablation study of grouping strategy. The best results are highlighted in bold.}
   \label{tab:abla_step}%
\end{table*}%

